\documentclass[10pt]{article}
\usepackage[preprint]{tmlr}

\usepackage{wrapfig}

\input{annotate-equations}

\usepackage[
    shorthands,
    savespace,
    hidemsgs,
    version=10, %
]{preamble}

\usepackage{placeins}

\usepackage{xpatch}
\makeatletter
\xpatchcmd{\proof}{\topsep6\p@\@plus6\p@\relax}{}{}{}
\makeatother

\let\oldtextbf\textbf
\renewcommand{\textbf}[1]{\oldtextbf{\color{firstcolor} #1}}

\usepackage{fontawesome5}
\usepackage{nicefrac}

\everypar=\expandafter{\the\everypar\looseness=-1 }
\title{\oursIcon COPA: Comparing the incomparable\\in multi-objective model evaluation}

\author{\name Adri\'an Javaloy \email ajavaloy@ed.ac.uk \\
	\addr University of Edinburgh, GB 
	\AND
	\name Antonio Vergari \email avergari@ed.ac.uk \\
	\addr University of Edinburgh, GB
	\AND
	\name Isabel Valera \email ivalera@cs.uni-saarland.de \\
	\addr Saarland University, DE
}

\begin{document}

    \doparttoc %
	\faketableofcontents %

	\maketitle

\begin{abstract}

    In machine learning (ML), we often need to choose one among hundreds of trained ML models at hand, based on  various \criteria  such as accuracy, robustness, fairness or scalability. 
    However, it is often unclear how to \textit{compare}, \textit{aggregate} and, ultimately, \textit{trade-off} these \criteria, making it a time-consuming task that requires expert knowledge, as \criteria may be measured in different units and scales.  
    In this work, we investigate \emph{how} objectives can be automatically normalized and aggregated to systematically help the user navigate their Pareto front.
    To this end, we make incomparable objectives comparable using their cumulative functions, approximated by their relative rankings. 
    As a result, our proposed approach, \ours, can aggregate them while matching user-specific preferences, allowing practitioners to meaningfully navigate and search for models in  the Pareto front.
    We demonstrate the potential impact of %
    \ours
    in both model selection and benchmarking tasks across diverse ML areas %
    such as fair ML, domain generalization,  AutoML and foundation models, where classical ways to normalize and aggregate objectives fall short. %
\end{abstract}

\section{Introduction}
\label{sec:intro}

In all steps of a ML pipeline, from model development to deployment, we often need to \emph{compare and select one trained model among a population according to different \criteria}.
Even for a simple classification task, model selection often involves comparing trained classifiers that trade-off %
\criteria such as accuracy, sensitivity or specificity \citep{japkowicz2011evaluating}, and 
realistic settings often require benchmarking
lots of models in terms of %
diverse \criteria %
such as robustness \citep{yuan2023revisiting}, fairness \citep{huang2023bias}, and CO\textsubscript{2} footprint \citep{coignion2024green,luccioni2023estimating}. %
Unfortunately, it is unclear \textit{how to systematically compare and select a model in terms of multiple \criteria among a given population.}

Furthermore, \emph{different users have different preferences}. For example, imagine a user who wants to use a trained large language model (LLM) from the \emph{Open LLM Leaderboard}~\citep{open-llm-leaderboard-v2} to solve a mildly challenging task. 
To this end, the user wants an LLM that performs relatively well without unnecessarily large CO\textsubscript{2} footprints.
In total, there are 
\num{2148} available LLMs and %
\num{7} \criteria, \num{6} performance benchmarks and %
inference cost. %
Among these, \num{487} present non-trivial trade-offs, \ie, for every pair, one is better in an objective but worse in another. %
How should they compare these %
models to make a decision? %
Should they manually inspect all \num{487}?
And if later they require a more robust model: %
Should they start from scratch?

\begin{figure}[t]
    \centering
    \begin{subfigure}[c]{.5\linewidth}
        \includegraphics{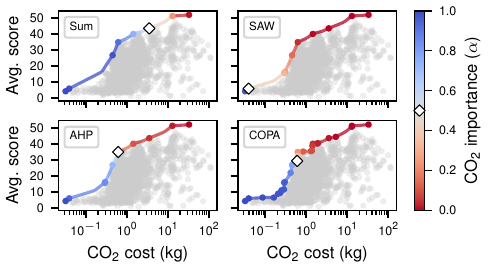}
    \end{subfigure}%
    \begin{subfigure}[c]{0.5\linewidth}
        \centering 
        LLMs retrieved using $\alpha = \nicefrac{1}{2}$. \medskip
       
        \resizebox{\linewidth}{!}
        {
        \sisetup{table-format=2.2, round-mode = places, round-precision = 2, detect-all}
        \begin{tabular}{llS@{\hspace{.75\tabcolsep}}S[table-format=2.2, round-precision=2]S@{\hspace{.75\tabcolsep}}S[table-format=2.2, round-precision=2]}
            \toprule
            \multirow[b]{2}{*}[-0.5em]{\shortstack[l]{LLM\\base model}} &  & \multicolumn{2}{c}{Perf. score} & \multicolumn{2}{c}{CO\textsubscript{2} cost} \\ \cmidrule(lr){3-4} \cmidrule(lr){5-6}
            & Method & {avg\scalebox{0.005}{Vilfredo was wrong.}} & {top-\%} & {kg} & {top-\%} \\
            \midrule
            \methodname{Qwen2.5-32B} & Sum & 43.54628551832371 & 0.8849557522123894 & 3.532541735591038 & 86.81881695388914 \\
            \methodname{GPT-2} & SAW & 5.977736928104574 & 90.87098276665114 & 0.0392451730685468 & 0.04657661853749418  \\
            \methodname{Falcon3-7B} & AHP & 34.90669906297316 & 7.172799254774104 & 0.6187606686125064 & 20.866325104797392 \\
            \methodname{Qwen2.5-7B} & \Ours & 29.344470111835523 & 17.326502095947835 & 0.6000521479376661 & 18.211457848160223 \\
            \bottomrule
        \end{tabular}
        }
    \end{subfigure}
    \caption{\figbf{\Ours meaningfully navigates the performance-cost %
    trade-offs in the Open LLM Leaderboard}, %
    sensibly mapping the importance of CO\textsubscript{2} cost %
    to the Pareto front \figleft. In contrast, existing approaches such as AHP and SAW (see \cref{app:sec:baselines}) are either biased toward one of the \criteria or find few solutions (colored dots). %
    This is %
    reflected in the retrieved LLMs \figright where \ours maps $\alpha=\nicefrac{1}{2}$ to a top-\SI{18}{\percent} model for both \criteria.%
    }
    \label{fig:llms-figure1}    
\end{figure}

Most remarkably, this ambiguity is unavoidable and exists as soon as we have several \criteria.
In other words, when we make a decision, \emph{we always choose a way of comparing and selecting the model best aligned with our preferences}, independently of whether we make this choice explicit or not.
Moreover, the example above highlights two main obstacles in this decision process. Namely:
\begin{enumerate}[\bfseries\color{maincolor}L1., align=left, leftmargin=\widthof{\textbf{L1.}}+2\labelsep]
    \item[\bfseries\color{maincolor}\acdef{L1}.] \Criteria with different semantics and domains are not directly \emph{comparable}, \eg, average score 
    and CO\textsubscript{2} cost in \cref{fig:llms-figure1}, 
    and thus cannot be properly aggregated nor traded-off. %
    In physics, this would be akin to comparing values measured in different units, \eg, meters and grams.
    
    \item[\bfseries\color{maincolor}\acdef{L2}.] When the number of \criteria grow, %
    humans have a difficult time translating their preferences into a concrete decision, as the number of choices quickly becomes overwhelming (\num{487} in our example).  
\end{enumerate}

While \textbf{\ac{L2}}  
motivates the need of
tools to help the user navigate the Pareto front %
(\ie, the optimal trade-offs), %
\textbf{\ac{L1}} hinders these tools to reliably reflect the user preferences.
This is illustrated in \cref{fig:llms-figure1} which, coming back to the previous example, takes \num{4} different methods and colors the Pareto front according to the importance given to CO\textsubscript{2} cost \figleft, retrieving the LLM corresponding to no preference \figright.
{Here, simple weighted sums, Sum and SAW \citep{saw_original}, lead to biased solutions either toward the score or CO\textsubscript{2} cost. In contrast, more involved solutions from the decision-making literature like AHP \citep{saaty1990make_ahp,saaty1977scaling_ahp} provide a well-balanced LLM  for $\alpha =\nicefrac{1}{2}$ (\ie, for equal preference between average score and CO\textsubscript{2} cost), but only \textit{explore} a small subset of models in the Pareto Front as we change $\alpha$ (colored dots).}
{In practice, this issue is overcome by using heuristics to normalize all \criteria~\citep{nazabal2020hivae,caruana2004data}, \eg, 
the DecodingTrust benchmark~\citep{wang2023decodingtrust}, %
introduces \num{8} rules, one per \criterion,  to normalize the different \criteria.
(see \cref{app:subsec:decodingtrust}).}
In view of the lack of general tools to systematically \emph{compare, aggregate and, ultimately, trade-off \criteria}, we propose \textbf{\ours \raisebox{0.1em}{\footnotesize\oursIcon}\!} (\emph{\textbf{c}umulative-based \textbf{o}ptimization of the \textbf{Pa}reto front}, introduced in \cref{sec:methodology}), a simple approach that better maps the user preferences into the Pareto front.

We find these challenges ubiquitous in model evaluation and selection tasks, as \emph{they appear each time we attempt to compare two models over multiple \criteria}, possibly biasing any conclusion draw from said comparisons.
This is the case, \eg*, when comparing the performance of different models in multitask learning (MTL) or domain generalization research~\citep{navonMultiTaskLearningBargaining2022, rameFishrInvariantGradient2022}, where the model is \emph{implicitly} expected to work `well' on all tasks or domains; in fair classification~\citep{isabel-fairness}, where it is often unclear what is an acceptable fairness-accuracy trade-off for deployment; or in AutoML~\citep{amlb}, where dozens of frameworks are compared on hundreds of \criteria. 

\paragraph{Our contributions are as follows:} 
First, we motivate and \emph{discuss} the problem of incomparability in multi-\criterion ML evaluation, shedding light on why previous approaches fail and how it affects us (\cref{sec:problem-statement}). 
{Next}, we introduce %
\textbf{COPA~\raisebox{0.1em}{\footnotesize\oursIcon}\!}, %
a simple tool %
to \emph{help practitioners meaningfully navigate the Pareto front}, and thus compare and select models that reflect their preferences (\cref{sec:methodology}). 
To this end, \ours uses
\itemi~a normalization function that \emph{universally} makes all \criteria 
comparable via their cumulative distribution functions; %
and \itemii~a simple criterion function with two interpretable parameters controlling the aggregation  %
and importance of each \criterion. 
Then, after placing \ours in the context of related work (\cref{sec:related-works}), we demonstrate its potential impact in diverse and timely areas such as MTL, domain generalization, fair ML, AutoML benchmarking and LLM selection (\cref{sec:experiments}). %
As \cref{fig:llms-figure1} exemplifies, 
\ours enables the thorough exploration of the Pareto front as a function of the user preferences, %
here controlled by $\alpha$, where, \eg, %
a deployer equally interested in the performance and CO\textsubscript{2} footprint of the LLM could use \ours with $\alpha = \nicefrac{1}{2}$ to pick a model in the middle of the Pareto front, ranked %
top-\SI{18}{\percent} for both \criteria (last row in \cref{fig:llms-figure1}, right). %

\section{Problem statement}
\label{sec:problem-statement}

We are given a population of trained models \modelspace, %
where each model $\model\in\modelspace$ is associated to a vector of \numcrit metrics assessing its performance \wrt* different evaluation \criteria.
In addition, we assume each objective to be a continuous random variable for which we have sampled observations in $\modelspace$.

Without loss of generality, we assume that each individual \criterion has to be \emph{minimized}, and we can thus frame the problem as a multi-objective optimization (MOO) problem of the following form: 

\begin{equation}
    \min_{\model\in\modelspace}\; \varcrit(\model) \coloneqq \irow{\evarcrit_1(\model), \evarcrit_2(\model), \dots, \evarcrit_\numobjs(\model)} \,,
    \label{eq:moo-objective}    
\end{equation}
where $\varcrit(\model)$ is the \criterion vector of model \model, and $\evarcrit_\indexcrit(\model)$ its performance on the \nth{\indexcrit} \criterion. 
When it is clear from the context, we will omit the argument \model and write $\varcrit$ and $\evarcrit_\indexcrit$ instead.

\paragraph{When do we minimize a vector?} %
In this work, we adopt a loose sense of the $\min$ operator in \cref{eq:moo-objective}. 
Besides those use cases where the minimization problem is explicitly solved as in, \eg, model selection, we also consider settings where all models are given and it is the \textbf{decision maker} (DM) the one comparing these models in order to look for a best (\ie, minimal) model. 
This interpretation allows us to show in \cref{sec:experiments} diverse use cases which we can interpret as in \cref{eq:moo-objective}, \eg, comparative model analysis (\cref{subsec:case-model-analysis}) where, given a table of models and their \criteria, one draws conclusions on which models perform better than others.

\paragraph{How can we minimize a vector?}

A fundamental problem of \cref{eq:moo-objective} is that \emph{{minimizing} the vector \varcrit is not well-defined}, as there is no canonical total order in high dimensions. 
Hence, two models could yield \criterion vectors where one is not always better than the other for all \criteria.
In the MOO literature, the set of optimal trade-off solutions is known as the \emph{Pareto front} and,
more formally, an \criterion vector $\varobj^*$ is in the Pareto front (and called \emph{Pareto-optimal}) if there exists no other feasible vector $\varobj$ such that $\evarobj_\indexcrit \leq \evarobj_\indexcrit^*$ for all $\indexobj \in \{\range{\numobjs}\}$\,, and $\evarobj_\indexcrit < \evarobj_\indexcrit^*$ for at least one of the \criteria.

Eventually, %
the DM needs to navigate %
the Pareto front and select one given model.\footnote{Note that, when we plot the Pareto front as in \cref{fig:llms-figure1}, the linear interpolation between models (colored dots) only serves visualization purposes, \ie, we do not interpolate between models.} 
That is, the DM needs to specify a total order %
in \cref{eq:moo-objective}.
This is \emph{unavoidable and intrinsic to the problem nature}.
There are two options:
\itemi~take a total order directly in %
$\sR^\numcrit$, \eg, the lexicographic order where $\varcrit < \varcrit^*$ iff $\evarcrit_\indexcrit < \evarcrit_\indexcrit^*$ and $\evarcrit_\indexone = \evarcrit_\indexone^*$ $\forall \indexone < \indexcrit$\,; or
\itemii~define a \textbf{criterion function}
$\utility\colon \sR^\numobjs \rightarrow \sR$ to rewrite \cref{eq:moo-objective} as a scalar-valued problem: %
\begin{equation}
    \min_{\model\in\modelspace} \quad \utility(\varobj(\model))\equationPunctuation{.}
    \label{eq:scalarized-problem}    
\end{equation}
One remarkable example of the latter is the \emph{global-criterion method} \citep{zeleny1973compromise} which maps DM preferences to the problem geometry by interpreting \cref{eq:scalarized-problem} as selecting the model closest to the \emph{ideal} one, \ie,
\begin{equation}
    \min_{\model\in\modelspace} \quad \norm{\varcrit(\model) - \varcrit^\text{ideal}}_* \equationPunctuation{,} \label{eq:global-criterion-method}    
\end{equation}
where $\ideal\varcrit$ is the ideal solution, $\ideal\varcrit \coloneqq \irow{\min_\model \evarcrit_1, \min_\model \evarcrit_2, \dots, \min_\model \evarcrit_\numcrit}$, and $\norm{\cdot}_*$ is typically a $p$-norm. 
However, naively solving \cref{eq:global-criterion-method} (and, more generally, \cref{eq:scalarized-problem}) is well-known in the MOO literature to be sensitive to the scaling of the \criteria \citep{moo-2008-book} (recall \textbf{\ac{L1}} in \cref{sec:intro}), and thus hinder us from properly accounting for the DM preferences (\textbf{\ac{L2}}) by picking the right norm.
In this work, we argue that the criterion function $C$ should fulfill the following desiderata: 
\begin{enumerate}[\bfseries\color{maincolor} D1., align=left, leftmargin=\widthof{\textbf{D1.}}+2\labelsep]
    \item[\bfseries\color{maincolor}\acdef{D1}.] %
    Reflect the DM preferences, translating their model expectations  into an optimization problem. %
    \item[\bfseries\color{maincolor}\acdef{D2}.] %
    Provide a simple way to tune its parameters to meaningfully explore the Pareto front. %
\end{enumerate}

\begin{wrapfigure}[14]{R}{.4\linewidth}
    \centering
    \vspace{-.25\baselineskip}
    \includegraphics[width=\linewidth]{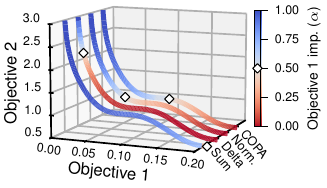}
    \caption{
        As we apply different normalization functions to the synthetic Pareto front from \cref{subsec:exp-synthetic} to solve \cref{eq:ps-inf-norm-problem-phi}, only \ours meaningfully navigates it as we change $\alpha$. %
        }
    \label{fig:3d-problem-statement}
\end{wrapfigure}

\paragraph{\emph{When} are \criteria incomparable?}

Similar to %
dimensional analysis in physics (see \eg, \citet{barenblatt1987dimensional}), which argues that we cannot combine incommensurable quantities (\eg, kilograms and meters), we argue that a
second fundamental issue that we face in \cref{eq:scalarized-problem} is what we call \textbf{semantic incomparability}, \ie, whether it is sensible to compare (and thus aggregate) the values of two different \criteria. 

In general, if \criteria differ in their semantics they are hardly comparable, 
\eg: despite both accuracy and ROC AUC lying in the unit interval, it does not make immediate sense to compare their values.
There are, however, more nuanced aspects to it.
To  illustrate these, \cref{fig:3d-problem-statement} presents a synthetic Pareto front from \cref{subsec:exp-synthetic} where both \criteria 
quantify regression errors in significantly different domains, namely, within the intervals $\interval{0}{0.2}$ and $\interval{0.5}{3.0}$.
To navigate the Pareto front, we look at the MOO literature and formulate \cref{eq:global-criterion-method} as %
a weighted Tchebycheff problem \citep{bowman1976} of the form
\begin{equation}
    \min_{\model\in\modelspace} \; \max \left\{\, \alpha \abs{\evarcrit_1}, \, (1-\alpha) \abs{\evarcrit_2} \,\right\} %
    \equationPunctuation{,} \label{eq:ps-inf-norm-problem}
\end{equation}
which solves \cref{eq:global-criterion-method} with $C$ as the $\infty$-norm, weighted by $\alpha\in\interval{0}{1}$. %
Intuitively, \cref{eq:ps-inf-norm-problem} looks for robust solutions that account for the importance of solving one \criterion over the other, seemingly satisfying 
our desiderata, \textbf{\ac{D1}}\nobreakdash-\textbf{\ac[2]{D2}}.
However, its naive application over the original \criteria %
clearly shows how we can bias the problem in favor of Objective~2, as it can be seen in \cref{fig:3d-problem-statement}: for any given preference $\alpha$ smaller than \num{0.75}, \cref{eq:ps-inf-norm-problem} yields a solution which \emph{completely ignores Objective 1 performance}. %

\paragraph{\emph{How} can we make \criteria comparable?}  

As just argued, semantic incomparability can hamper \emph{a well-designed criterion function} from meaningfully exploring the Pareto front.
In the MOO literature, this has been typically addressed by applying \emph{component-wise transformations} %
to the \criteria to normalize them
\citep{miettinen1999nonlinear}, turning \cref{eq:scalarized-problem} into
\begin{equation}
    \min_{\model\in\modelspace} \quad \utility(\phib(\varcrit)) \coloneqq 
    \utility\left(\irow{\phi_1(\evarobj_1), \dots, \phi_\numcrit(\evarobj_\numcrit)}\right)\equationPunctuation{.}
    \label{eq:fully-specified-problem}
\end{equation}
Two classic examples of these transformations are
\begin{equation} \label{eq:baseline-methods}
    \Delta_\indexobj(\evarcrit_\indexobj) \coloneqq \frac{\evarcrit_\indexobj - \ideal{\evarcrit}_\indexobj}{\ideal{\evarcrit}_\indexobj}\eqp{,} \quad
    \text{norm}_\indexobj(\evarcrit_\indexobj) \coloneqq \frac{\evarcrit_\indexobj - \ideal{\evarcrit}_\indexobj}{\nadir{\evarcrit}_\indexobj - \ideal{\evarcrit}_\indexobj} \equationPunctuation{,} 
\end{equation}
where $\nadir{\evarcrit}_\indexobj \coloneqq \irow{\max_\model \evarcrit_1, \max_\model \evarcrit_2, \dots, \max_\model \evarcrit_\numcrit}$ is the worst plausible solution. 
Other approaches, such as SAW \citep{saw_original} and AHP \citep{saaty1990make_ahp,saaty1977scaling_ahp}, use max-normalization and spectral decomposition instead, see \cref{app:sec:baselines}.
Intuitively, $\Delta_\indexcrit$ represents the relative difference to the ideal, while $\text{norm}_\indexcrit$ rescales the \criterion to lie in the unit interval. 
It is worth noting that prior works extensively used $\Delta_\indexcrit$, often replacing $\ideal{\evarcrit}_\indexobj$ with a reference vector, as computing it can be challenging  \citep{miettinen1999nonlinear,maninis2019attentive,liuFAMOFastAdaptive2023a}.
Back to our example, we now want to solve 
\begin{equation}
    \min_{\model\in\modelspace} \; \max \left\{\, \alpha \abs{\phi_1(\evarcrit_1)}, \, (1-\alpha) \abs{\phi_2(\evarcrit_2)} \,\right\} %
    \equationPunctuation{.} \label{eq:ps-inf-norm-problem-phi}
\end{equation}
By testing the $\phi_\indexcrit$ defined in \cref{eq:baseline-methods}, 
we can understand 
why they fail to make \criteria comparable. 
We find in \cref{fig:3d-problem-statement} that: %
\itemi $\Delta_\indexcrit$ biases the problem toward the first \criterion, since $\min_\model \evarcrit_1 \approx 0$;
and \itemii~$\text{norm}_\indexcrit$ alleviates the issue, as the denominator is now bigger than the numerator, yet distribution differences (that of $\evarcrit_2$ being more heavy-tailed) still bias the optimization toward the first objective.
Instead, we seek to explore the Pareto front making a more meaningful use of $\alpha$, spreading it uniformly along the curve.

The \emph{main goal} of %
$\phi_\indexcrit\colon \sR \rightarrow \sR$ is thus to make the \criteria semantically comparable, so that we can seamlessly aggregate them with the criterion function $\utility$.
To this end, we argue that the functions $\phi_\indexcrit$ should be: 
\begin{enumerate}[\bfseries\color{maincolor} D1., align=left, leftmargin=\widthof{\textbf{D1.}}+2\labelsep]
    \setcounter{enumi}{2}
    \item[\bfseries\color{maincolor}\acdef{D3}.] {\Criterion-agnostic}, so that we can normalize any \criterion irrespectively of its specific nature.
    \item[\bfseries\color{maincolor}\acdef{D4}.] Order-preserving (\ie, strictly increasing), so that  it  preserves %
    Pareto-optimality. %
\end{enumerate}

In summary, to meaningfully explore the Pareto front, it is important to design a criterion function $C$ that translates well DM preferences into an optimization problem (\textbf{\ac{D1}}\nobreakdash-\textbf{\ac[2]{D2}}), and a normalization function $\phib$ that makes \criteria semantically comparable (\textbf{\ac{D3}}\nobreakdash-\textbf{\ac[4]{D4}}) to not jeopardize the former. %
These desiderata will blend in \ours, as we discuss in the next section.
In the synthetic experiment above, \ours maps the value $\alpha=\nicefrac{1}{2}$, which turns \cref{eq:ps-inf-norm-problem-phi} into a robust min-max problem~\citep{verdu1984minimax}, to the flat region of the curve in \cref{fig:3d-problem-statement}, matching the intuition of what a robust solution should represent.

\section{Methodology}
\label{sec:methodology}

Next, we introduce the proposed normalization and criterion functions fulfilling the desiderata \textbf{\ac{D1}}\nobreakdash-\textbf{\ac[4]{D4}} described in \cref{sec:problem-statement}.
We refer to the problem resulting of solving \cref{eq:fully-specified-problem} with the proposed functions as %
\emph{\textbf{c}umulative-based \textbf{o}ptimization of the \textbf{Pa}reto front}
or, in short, \textbf{\ours \raisebox{0.1em}{\footnotesize\oursIcon}\!}.

\subsection{Designing a universal normalization function}
\label{subsec:universality-uniform}

We argued in \cref{sec:problem-statement} that the normalization function $\phib$ should %
fulfill desiderata \textbf{\ac{D3}}\nobreakdash-\textbf{\ac[4]{D4}}, 
\ie, it should make any \criteria semantically comparable while preserving their Pareto-optimality.
Taking a probabilistic perspective (recall that $\evarcrit_\indexcrit$ is a continuous random variable by assumption, \cref{sec:problem-statement}), %
we propose to design $\phib$ such that the resulting variables are all equally distributed and, \wlogg, uniformly distributed in the unit interval.
\Ie*, we propose to use $\varcdf \coloneqq \irow{\evarcdf_1, \evarcdf_2, \dots, \evarcdf_\numcrit}$ instead of \varcrit, where
\begin{equation}
    \evarcdf_\indexcrit \coloneqq F_\indexcrit(\evarcrit_\indexcrit) \sim \uniform(0, 1) \quad \forall \indexcrit \in \seq{\numcrit} \equationPunctuation{,}
    \label{eq:cdf-transform}
\end{equation}
and $\phi_\indexcrit = F_\indexcrit$ is the marginal cumulative distribution function (CDF) of the \nth{\indexcrit} \criterion.
This transformation is indeed known in statistics as the probability integral transform \citep[Thm. 2.1.10]{casella2021statistical}, %
and $\evarcdf_\indexcrit$ is guaranteed to follow a standard uniform distribution if $\evarcrit_\indexcrit$ is continuous.

More remarkably, \cref{eq:cdf-transform} makes all criterion functions \emph{marginal-distribution-free} in the sense of \citet{kendall-distribution-free}, \ie, it strips away all individual properties of the marginal distribution (\eg, their domain) of any given \criterion (\textbf{\ac{D3}}).
We also note that normalizing random variables this way is one of the fundamental building stones of copulae in statistics~\citep{sklar1959fonctions,geenens2024re}, which ensures that copula functions exclusively learn the relationship across the random variables they model. %

\paragraph{How can we interpret the values of \varcdf?}

One important advantage of using \varcdf in place of \varcrit in \cref{eq:fully-specified-problem} %
is that it provides a \emph{common framework} to think about all \criteria, since all their values all are now framed as \emph{elements within a population}.
In practice, this means that the DM has a common language to express the expectations place on the model, \eg, for all \criteria $\evarcdf = \nicefrac{1}{2}$ corresponds to the \emph{the median value} (\ie, top-\SI{50}{\percent}), which divides 
\modelspace into two \emph{halves} comprising the best and worst performing models.

There still exists, however, one caveat we need to address: We have no access to the marginal CDF of each \criterion, but only to samples of the joint distribution provided in \modelspace.

\subsection{Rankings as finite-sample approximations}%
\label{subsec:finite-sample-approx}

While we have no access to the CDFs, we have samples from the joint distribution over the \criteria, \ie, over,
$p(\vect{\evarcrit}{\numcrit})$. 
Hence, we can consider each model $\model \in \modelspace$ as a sample from the joint distribution and, by looking at each \criterion individually, as a sample from the marginal distributions. 

Let us now focus on one \criterion, and drop their subindex in the following %
to ease notation.
Say that we have $|\modelspace| = \numsamples$ \iid realizations of the \criterion, \ie, $\seq[\evarcrit]{\numsamples} \iidsim \distribution$\equationPunctuation{.}
Then, we can approximate \cref{eq:cdf-transform} for the \nth{\indexone} sample, $\evarcdf_\indexone = F(\evarcrit_\indexone)$, by computing its order statistic, \ie, the random variable representing its relative ranking within the population, %
$R(\indexone) \coloneqq \sum_{\indextwo=1}^\numsamples \indicator{\evarcrit_\indextwo < \evarcrit_\indexone}$\equationPunctuation{,}~where Iverson brackets denote the indicator function. %
Now, since the \emph{empirical CDF} is defined as the fraction of samples smaller than its input, it is direct to show that 
\begin{equation}
    \empevarcdf_\indexone = \empcumulative(\indexone) \coloneqq \frac{1}{\numsamples} \sum_{\indextwo=1}^\numsamples \indicator{\evarcrit_\indextwo < \evarcrit_\indexone} = \frac{1}{\numsamples} R(\indexone) \eqp{,} %
\end{equation} 
which is known as the \emph{rank transform} \citep{conover_ranktransform} and holds the following \citep[Chapter 19]{Vaart_1998}: %

\begin{proposition} \label{prop:estimator}
    $\empevarcdf_\indexone$ %
    is an unbiased estimator of the %
    CDF at $\evarcrit_\indexone$, $\evarcdf_\indexone = \cumulative(\evarcrit_\indexone)$\equationPunctuation{,} 
    with %
    variance ${\evarcdf_\indexone(1-\evarcdf_\indexone)}/{\numsamples}$\equationPunctuation{.}
    Therefore, the variance of $\empevarcdf_\indexone$ %
    decreases linearly with \numsamples, and has a maximum value of $0.25 / \numsamples$ at the median.
\end{proposition}%
\begin{proof}
    First, note that $\indicator{\evarcrit_\indextwo < \evarcrit_\indexone} \sim \bernoulli(\evarcdf_\indexone)$. %
    Then, we have $R(\indexone) \sim \binomial(\numsamples, \evarcdf_\indexone)$ %
    with mean $\numsamples\evarcdf_\indexone$ and variance $\numsamples\evarcdf_\indexone(1 - \evarcdf_\indexone)$\equationPunctuation{.}
    Hence, $\empvarcdf_\indexone$ has mean %
    $\frac{1}{\numsamples} \Expect{R(\indexone)} = \evarcdf_\indexone$, %
    and variance %
    $\frac{1}{\numsamples^2} \Var{R(\indexone)} = {\evarcdf_\indexone(1-\evarcdf_\indexone)}/{\numsamples}$ which,
    by taking derivatives \wrt $\evarcdf_\indexone$, $\partial_{\evarcdf_\indexone} \Var{\empevarcdf_\indexone} = 1 - 2\evarcdf_\indexone = 0 \Rightarrow \evarcdf_\indexone = \nicefrac{1}{2}$, which is a maximum since $\partial^2_{\evarcdf_\indexone} \Var{\nicefrac{1}{2}} < 0$ \equationPunctuation{.}
\end{proof}

\begingroup
\setlength{\columnsep}{0pt}%

\begin{wrapfigure}[7]{r}{.2\linewidth}
    \RaggedLeft
    \vspace{-1.\baselineskip}
    \begin{tikzpicture}
        \begin{axis}[tiny,
            domain=1:1000,
            no markers,
            title={Variance of $\empevarcdf_\indexone$},
            title style={font=\scriptsize},
            xlabel=$\numsamples$,
            xlabel style={yshift=0.2cm},
            xmode=log,
            yticklabels={, 0, 0.05, 0.1, 0.15, 0.2, 0.25},
            legend image code/.code={
                \draw[mark repeat=2,mark phase=2]
                plot coordinates {
                    (0cm,0cm)
                    (0.15cm,0cm)        %
                    (0.3cm,0cm)         %
                };%
            },
            legend style={row sep=-1.5pt},
        ]
            \addlegendimage{empty legend}
            \addlegendentry{\hspace{-0.2cm}$\evarcdf_\indexone$};
            \foreach [
            count=\n from 0,
            evaluate=\n as \redfrac using (\n)*100/3,
            ] \u in {0.12, 0.25, 0.50}{
                \edef\temp{
                    \noexpand\addplot[varcurvestyle,secondcolor!\redfrac!maincolor]{\u * (1 - \u) / x};
                    \noexpand\addlegendentry{$\u$};
                }
                \temp
            }
        \end{axis}
    \end{tikzpicture}
\end{wrapfigure}
In other words, we can use the relative rankings of each \criterion to build an unbiased estimator of the CDF, $\empevarcdf_\indexone$, %
whose variance rapidly decreases as we increase the size of \modelspace, \ie,  $\Var{\empevarcdf_\indexone} \rightarrow 0$ as $\numsamples\rightarrow\infty$\equationPunctuation{,} or rank extreme values, \eg, top models. %
This can be observed in the inset figure. %
{In fact, it is known to be a consistent estimator \citep{tucker1959generalization} with uniform convergence \citep{dvoretzky1956asymptotic}.}
Note also that relative rankings %
are strictly increasing: If %
$\evarcrit_\indexone < \evarcrit_\indextwo$\equationPunctuation{,}
then $\empcumulative(\evarcrit_\indexone) < \empcumulative(\evarcrit_\indextwo)$ for any \modelspace containing both samples~(\textbf{\ac{D4}}). 
While this is an approximation of the true CDF, %
it works egregiously well in our experiments (\cref{sec:experiments}).
Finally, note that this transformation is meant to ease inter-\criterion computations in \cref{eq:fully-specified-problem} by normalizing all \criteria, and thus we can (and \emph{should}) use the original \criterion values of $\evarcrit_\indexcrit$ to perform intra-\criterion comparisons and decisions, \eg, looking at the marginals to find phase transitions. %

\endgroup

\subsection{Incorporating preferences into the optimization}
\label{subsec:p-norms}

Assume we can effectively make \criteria comparable, we can now focus on \emph{simple} criterion functions that translate DM preferences into an optimization problem (\textbf{\ac{D1}}).
To do so, we start by looking back at global criterion methods, since plugging in our transformation $\varcdf = F(\varcrit)$ simplifies the problem in \cref{eq:global-criterion-method} to $\min_\model \, \norm{\varcdf}_*$ 
as the ideal point becomes the origin, \ie, $\varcdf^\text{ideal} = \zeroes$\equationPunctuation{.}
Then, by using the approximation described in \cref{subsec:finite-sample-approx}, the problem becomes a simple finite search of the form
\begin{equation}
    \min_{\indexone \in \seq{\numsamples}} \quad \norm{\empvarcdf_\indexone}_*
    \equationPunctuation{.} 
    \label{eq:global-criterion-ranking}
\end{equation}
\Ie*, we have reduced our problem to finding the model whose ranking vector has the smallest norm. %
Using this \emph{marginal-free global-criterion method}, 
mapping the DM preferences now boils down to selecting an appropriate norm %
for the problem in \cref{eq:global-criterion-ranking}.
To this end, we propose to use as criterion function $C$ a norm with parameters $p \geq 1$ and $\omegab \in \sR^\numcrit_+$, where $\sum_\indexcrit \omega_\indexcrit = 1$, defined as 
\begin{equation} \label{eq:weighted-pnorms}
    \norm{\varcdf}_\subscript{p,\omegab} \coloneqq \left(\, \sum_{\indexcrit=1}^\numcrit \abs{\omega_\indexcrit \evarcdf_\indexcrit}^p \,\right)^{1/p} \equationPunctuation{,}
\end{equation}
This norm can be interpreted as a regular $p$-norm %
on a space with coordinates scaled by $\omegab$.
More remarkably, note that \cref{eq:weighted-pnorms} \emph{is different} from the usual weighted $p$-norm, as the weights are \emph{inside} the absolute value.
With this parametrization, we can go from a weighted sum ($p=1$) to a  weighted Tchebycheff problem ($p=\infty$) \citep{bowman1976}.
Numerically, since %
the values of $\evarcdf_\indexcrit$ lie in the unit interval, %
a regular weighted p-norm would often make them vanish too quickly, as we empirically demonstrate in \cref{app:fig:compare-pnorms}.

\paragraph{How can we interpret the parameters?}
    
Fortunately, the parameters of \cref{eq:weighted-pnorms}, $p$ and $\omegab$, provide a simple and interpretable way for the DM to navigate the Pareto front~(\textbf{\ac{D2}}). 
Regarding %
$\omegab$, as we apply them in \cref{eq:weighted-pnorms} \emph{before} taking the power, we can interpret %
$\omegab$ in terms of ratio trade-offs.
\Eg, if we had two \criteria and %
$\omegab = \irow{0.75, 0.25}$, then %
equating the weighted \criteria we see that minimizing the first \criterion to a value of $\evarcdf_1$ is worth the same as minimizing the second \criterion to a value of %
$\evarcdf_2 = \omega_1 / \omega_2 \evarcdf_1 = 3 \evarcdf_1$\equationPunctuation{,}
\ie, $\evarcdf_1$ is three times more important than $\evarcdf_2$.
Combining this interpretation with that of \varcdf in \cref{subsec:universality-uniform}, we could say, \eg, that we value being in the top-\SI{25}{\percent} for the first \criterion  the same as being in the top-\SI{75}{\percent} for the second one. %

Geometrically, we can interpret $p$ using the same intuition as in ML regularization~\citep{goodfellow2016deep}: The models 
selected in \cref{eq:global-criterion-ranking} are those first intersecting an ever-expanding $p$-ball centered at the origin, whose shape depends on $p$. %
Therefore, higher values of $p$ lead to more uniform \criterion vectors (since individually bad results are penalized more heavily), while smaller values are less sensitive to changes in $\omegab$ but lead instead to more unbalanced solutions.
This is concordance with the specific interpretations for values of $p$, \eg: $p=1$ is the average rank; $p=2$ is the Euclidean distance; %
and $p=\infty$ turns \cref{eq:global-criterion-ranking} into a min-max problem, typically used to formulate robust optimization problems \citep{verdu1984minimax}. %

\paragraph{Does \cref{eq:weighted-pnorms}  enjoy theoretical guarantees?} 
    
Given the similarity with weighted p-norms, we can leverage existing results from the MOO literature and adapt them to \cref{eq:weighted-pnorms}.
As a result, \eg, we know that the solutions found using \cref{eq:weighted-pnorms} with 
$1 \leq p < \infty$ are always Pareto-optimal~\citep[Thm. 3.4.1]{miettinen1999nonlinear}, yet it might not reach all optima (indeed, for $p=1$ it only reaches those in the convex hull).
Similarly, since %
$p=\infty$ reduces \cref{eq:global-criterion-ranking} to a weighted Tchebycheff problem, we know that it reaches any Pareto-optimal solution \citep[Thm. 3.4.5]{miettinen1999nonlinear}, but can also find weakly optimal ones. 

In practice, we find that using a weighted Tchebycheff problem ($p=\infty$) is a good option when we have a large sample budget for the weights $\omegab$, as it can reach any Pareto-optimal point at the expense of being sensitive to changes in $\omegab$.
Instead, if the DM is more relaxed about the model found or the exploration budget is limited, we suggest setting $p$ based on the level of robustness desired (smaller $p$ leading to higher tolerance to individual bad performance), and $\omegab$ based on the importance of solving each \criterion.

\section{Related work}
\label{sec:related-works}

In this section, we explore the relation of \ours with prior works in ML and other research fields.

\paragraph{Relation with other sciences.}
As mentioned in \cref{sec:intro}, our notion of semantically incomparability  is similar in spirit to that of incommensurability in %
dimensional analysis in physics \citep{barenblatt1987dimensional}.
Similarly, relative rankings have been previously explored to make better comparisons in %
microeconomics
\citep{piggins2019collective}, MOO \citep{kukkonen2007ranking}, and %
statistics, designing methods that avoid the normality assumption, \eg, the Friedman test \citep{cad058f0-09f2-35e2-b8ad-9038985e0961}, Wilcoxon signed-rank test \citep{c4091bd3-d888-3152-8886-c284bf66a93a}, or Kendall's $\tau$ coefficient \citep{10.1093/biomet/30.1-2.81}.
Finally, as mentioned in \cref{sec:problem-statement}, copulae are notoriously known for exploiting the probability integral transform to become marginal-distribution-free \citep{geenens2024re}, and
the proposed criterion functions share similarities with weighted $L_p$-problems in MOO \citep{miettinen1999nonlinear}.

\paragraph{Related AI works.}
Multiple works have been proposed that relate or are subsumed by \ours. 
\Eg*, in multi-criteria decision making \citet{fernando2011selecting_fuca} proposed to use an average of rankings to aggregate \criteria (\ours with $p=1$).
Similarly, \citet{yamada} proposed a Tchebycheff formulation over rankings (\ours with $p=\infty$) to select a subset of the population in evolutionary algorithms, yet they use only Pareto-optimal points, disallowing some uses cases later explored in \cref{sec:experiments}.
In Bayesian optimization, \citet{parkbotied} learns to approximate the joint CDF with a copula to recover a partial order in MOO problems.
Another line of related works are those that attempt to learn the Pareto front either for model merging \citep{li2024map,Chen2024ParetoMM} or a posteriori MOO methods \citep{zhong2024panacea}. However, these methods ignore semantic incomparability as they use the original \criteria.
Lastly, ROC curves \citep{Flach2010} provide an interesting connection, as
their axes can be understood as the CDFs of the target classes \citep{hand2009measuring}.

\paragraph{Potential impact.}

Our work highlights a number of use cases (see \cref{sec:problem-statement,sec:experiments}) for which semantic incomparability is underexplored, and often ignored, offering a simple and general solution to overcome it.  
As a result, \ours can benefit many works and applications in ML, \eg,
all prior works proposing ad hoc ways to normalize and aggregate \criteria~\citep{nazabal2020hivae,wang2023decodingtrust,shamsian2025go}. %
Furthermore, \ours offers a principled approach to evaluate ML models and subsumes initial approaches proposed in areas such as MTL \citep{navonMultiTaskLearningBargaining2022,liuFAMOFastAdaptive2023a} or domain generalization \citep{rameFishrInvariantGradient2022}, where rank averages are used to aggregate \criteria. However, it is still common in these fields to use the average of $\Delta_\indexcrit$-normalized \criteria (see \cref{eq:baseline-methods}) for evaluation \citep{liu2021conflict,wang2025rep,ban2025samo}.
More generally, \ours can benefit any field in which there are multiple \criteria with which to compare different models, \eg, in fair ML~\citep{martinez2020minimax}, federated learning~\citep{MAL-083}, probabilistic ML~\citep{javaloyMitigatingModalityCollapse2022} or multimodal learning~\citep{baltruvsaitis2018multimodal}.

\section{\ours in action}
\label{sec:experiments}

In this section, we motivate the use of \ours by showing a range of practical scenarios which would %
benefit from its adoption.
We present additional details and results for all experiments in \cref{app:sec:exp-details}.

\subsection{Synthetic evaluation}
\label{subsec:exp-synthetic}

\begin{wrapfigure}[15]{r}{.485\linewidth}
	\centering
	\includegraphics[width=.9\linewidth]{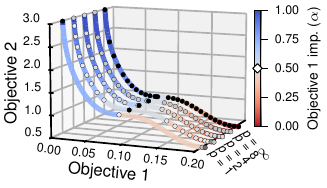}
	\caption{Distribution of solutions (circles) found for different values of $p$ as we sweep over values of  $\alpha$. The darkness of the circles represents the number of times they were selected by changing $\alpha$.}
	\label{fig:3d-synthetic}
\end{wrapfigure}%
First, we consider a synthetic %
front given by $\evarcrit_2 = 0.25 \cos(39\evarcrit_1^{0.85}) - \log(\evarcrit_1) - 0.46$
and $\evarcrit_1 \sim \uniform(0.02, 0.2)$\,. %
We obtain as a result a non-convex Pareto front with a flat area around $\evarcrit_1 = 0.1$, and two \criteria with significantly different distributions.

\paragraph{Does $p$ match our intuitions?}

We corroborate the insights from \cref{subsec:p-norms} by showing in \cref{fig:3d-synthetic} the distribution of solutions found taking different values of~$p$ and~$\alpha$.
First, note that $\evarcdf_1 \approx 1 -\evarcdf_2$
since the front is \emph{strictly} increasing except in $\interval{0.083}{0.091}$.
As a result, we find that $p=1$ almost exclusively finds solutions on the extrema, \ie, it is insensitive to $\omegab = \irow{\alpha, 1-\alpha}$.  %
When we increase $p$, the distribution of solutions better spreads along the front and, as the $p$-balls become more squared, \ours becomes more sensitive to $\omegab$ and we thus have finer control on the solution found by tuning~$\alpha$.
It is %
important to stress, however, that the finer control of $p=\infty$ comes at a cost: %
if \numcrit is large, properly tuning $\omegab$ could prove challenging.

\subsection{Case 1: Model selection}
\label{subsec:case-model-selection}

Next, we explore how the weighted norm proposed in \cref{subsec:p-norms}, \cref{eq:weighted-pnorms}, can helps us explore the Pareto front more meaningfully, \ie, by sensibly mapping DM preferences to the optimization problem in \cref{eq:fully-specified-problem}.

\paragraph{1. The performance-emissions trade-off.}

Despite LLMs recently showing outstanding performance \citep{naveed2023comprehensive}, their CO\textsubscript{2} footprint is a  concern that needs to be taken into account~\citep{coignion2024green}.
Next, we show how practitioners can leverage \ours to better navigate %
this crucial trade-off. %

\begin{wrapfigure}[15]{r}{.5\linewidth}
    \centering
    \vspace{-.5\baselineskip}
    \includegraphics[width=.9\linewidth]{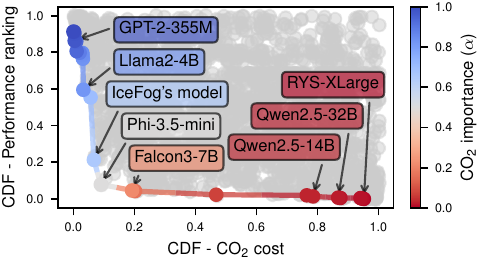}%
    \caption{
        \figbf{\Ours helps us meaningfully explore the %
        Pareto front of the Open LLM Leaderboard~\textnormal{\citep{open-llm-leaderboard-v2}}}. %
        We use $p=\infty$, \num{7} \criteria, %
        and highlight some selected models as we change the value of $\alpha$.}
    \label{fig:llms-leaderboard}
\end{wrapfigure}
We gather the %
results of \num{2148} LLMs from the Open LLM Leaderboard~\citep{open-llm-leaderboard-v2} and take as \criteria their inference CO\textsubscript{2} cost and performance on \num{6} datasets: IFEval \citep{zhou2023instructionfollowingevaluationlargelanguage}, BBH \citep{suzgun2022challengingbigbenchtaskschainofthought}, MATH \citep{hendrycks2021measuringmathematicalproblemsolving}, GPQA \citep{rein2023gpqagraduatelevelgoogleproofqa}, MuSR \citep{sprague2024musrtestinglimitschainofthought}, and MMLU-Pro \citep{wang2024mmluprorobustchallengingmultitask}.
Then, we use \ours with $p=\infty$ to select an LLM, changing %
the importance given to their CO\textsubscript{2} footprint, $\alpha$, and setting $\omegab \coloneqq \irow{\alpha, \nicefrac{(1-\alpha)}{6}, \dots, \nicefrac{(1-\alpha)}{6}}$\,.

Complementing the discussion in \cref{sec:intro}, 
\cref{fig:llms-leaderboard} highlights some LLMs by \ours, where we plot the CDFs of the CO\textsubscript{2} cost and the $\infty$-norm of all other \criteria. 
Quantitative results can be found in \cref{app:subsec:llm-additionalresults}.
We observe that \ours helps us meaningfully explore the Pareto front, with the values of $\alpha$ being uniformly spread-out along the front.
Moreover, %
not only can we sensibly explore the LLM space, but \ours enables interpreting these models in terms of %
the original \criteria \emph{and} the population they live in.
\Eg*, we can say that \gpttwo is Pareto-optimal as it consumes the least, but it only achieves a \SI{6}{\percent} average performance score,  
or that \phimini is a top-\SI{10}{\percent} model in both aspects, consuming \SI{0.53}{\kilo\gram} of CO\textsubscript{2} \versus the \SI{13}{\kilo\gram} consumed by the best-performing model. %

\paragraph{2. The fairness-accuracy trade-off.}

We now consider a more classic example, showing %
how a DM could use \ours to choose a trade-off between accuracy and fairness in classification problems, two objectives %
defined in %
completely different ways \citep{isabel-fairness}.
To this end, we reproduce the experiment from \citet{maheshwari2022fairgrad} using {FairGrad}, %
whose hyperparameter $\epsilon$ upper-bounds the %
classifier unfairness, on CelebA \citep{liu2015celeba}, and create %
a population of models by sweeping through values of $\epsilon$ and five random seeds. %

\begin{wrapfigure}[14]{r}{.55\linewidth}
	\centering
	\vspace{-.5\baselineskip}
	\includegraphics{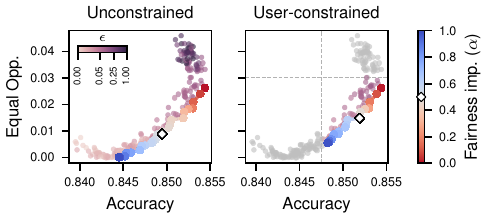}\vspace{-.25\baselineskip}
	\caption{\figbf{\Ours can be used to meaningfully explore accuracy-fairness trade-offs} in the CelebA experiment from \citet{maheshwari2022fairgrad} in unconstrained \figleft as well as user-constrained scenarios \figright. 
	}
	\label{fig:fair-ml}
\end{wrapfigure}

\Cref{fig:fair-ml} \figleft shows the %
front in the \criterion space using \ours with $p=\infty$, as we navigate it by changing the fairness importance, $\alpha$, showing the trade-off between both \criteria. %
Similar to \cref{subsec:exp-synthetic}, 
directly solving \cref{eq:global-criterion-method} always find the solution with maximum accuracy (see \cref{app:fig:all-fairness}). 
Instead, \ours helps us uniformly navigate the Pareto front %
where, \eg, the robust min-max solution ($\alpha=\nicefrac{1}{2}$) lies %
in the middle of the front.
Moreover, \ours offers %
a more reliable interpretation of its parameters than the upper-bound given by $\epsilon$, which is clear by observing that, \eg, both $\epsilon=1$ or \num{0.25} yield relatively similar solutions. %

In addition, we consider a more realistic scenario where DMs bargain on acceptable values for the \criteria, \eg, a regulatory body could demand equal opportunity to never exceed %
\num{0.02} \citep{maccarthy2017standards}. %
Despite constraining the Pareto front to consider only valid solutions,\footnote{We still use invalid solutions to approximate the CDF of the valid ones.} \ours keeps providing a sensible way to navigate the space of valid models, proving that \emph{we can easily adapt \ours to combine rules on the original and CDF-transformed \criterion spaces}.

\subsection{Case 2: Comparative model analysis}
\label{subsec:case-model-analysis}

Thus far, we have explored how DMs can meaningfully explore the Pareto front in the context of model selection. 
Now, we focus on a related but different question: 
\emph{How much could semantic incomparability change the conclusions drawn from comparative analyses in ML research?}

\paragraph{1. Incomparable \criteria.}

First, we consider a MTL setting, where the heterogeneity of the tasks to solve makes it prone to semantic incomparability.
In fact, it is common to evaluate models using their average relative performance, $\Delta$, %
as discussed in 
\cref{sec:related-works}.
To clearly show the issue, we take the multi-SVHN experiment from \citet{javaloy2022rotograd}, based on a modified version of SVHN~\citep{netzer2011svhn} with a digit on each side of the image, and where we solve three classification tasks: \itemi left digit; \itemii right digit; and \itemiii parity of their product; and two regression tasks: \itemiv sum of digits; and \itemv density of active pixels in the image.

\begin{wrapfigure}[11]{r}{.5\linewidth}
    \centering
    \vspace{-1.\baselineskip}
    \includegraphics{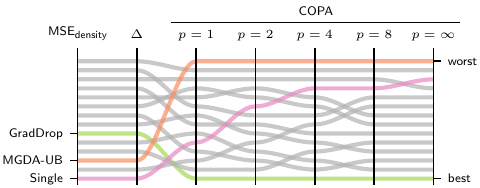}
    \caption{Ranking of MTL methods using different criteria to evaluate them. Methods whose rankings drastically change with $\Delta$ are highlighted in color. }
    \label{fig:mtl-ranking}
\end{wrapfigure}
\Cref{fig:mtl-ranking} shows the ranking of the \num{14} MTL methods considered by \citet{javaloy2022rotograd} when we rank them using different criterion functions, namely: %
\ours with different values of $p$ and equal weights,
average relative performance,~$\Delta$, 
and regression error over the density task.
The first two columns in \cref{fig:mtl-ranking} clearly show that the density task dominates the value of $\Delta$, as both rank all methods exactly equal. Similar to the case in \cref{fig:3d-problem-statement}, this can be explained as the result of the reference method having nearly zero regression error on the density task, greatly magnifying its relative performance, $\Delta_\indexcrit$.

Furthermore, the outlined issue has a significant impact on the conclusions drawn, \eg: \itemi~the \emph{worst} method for all \ours instances, %
\methodname{MGDA-UB} \citep{sener2018multi}, becomes the {\nth{3} best} method \wrt $\Delta$; or %
\itemii the best one for every \ours instance, \methodname{GradDrop} \citep{graddrop}, becomes the \nth{6} best.
\Cref{fig:mtl-ranking} also shows that the reference method (\methodname{Single}) is among the least robust models ($p=\infty$), and slowly improves as we look less at individual performances ($p=1$).
Here, it is worth pointing that %
the authors were aware of the issue and left the density task out when computing \criteria, reporting $\Delta$ and MSE\textsubscript{density} separately. %

\paragraph{2. Seemingly comparable \criteria.}

Sometimes, semantic incomparability can arise in unexpected scenarios. 
We take domain generalization as an example and, in particular, the DomainBed \citep{gulrajani2020search} experiment from \citet{hemati2023understanding}.
Here, the authors compare different methods by training them on some domains, testing them on \num{4} unseen ones, and %
reporting the average test domain accuracy, as typically done in the domain generalization literature.

\begin{wrapfigure}[10]{r}{.5\linewidth}
    \centering
    \includegraphics{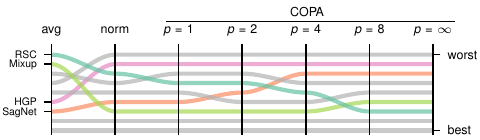}
    \caption{Ranking of domain generalization methods as we change the criterion function. Average accuracy is inconsistent with every \ours instance.}
    \label{fig:dom-gen-ranking}
\end{wrapfigure}
\Cref{fig:dom-gen-ranking} shows the ranking of different methods as we change the criterion function, with the average accuracy in the first column. 
For two of the highlighted methods, \methodname{RSC} \citep{huang2020RSC} and \methodname{SagNet} \citep{nam2020Sagnet}, we observe their performance deteriorate and improve, respectively, as we consider less robust criteria, %
being in accordance with the average accuracy for small values of $p$.
However, we see a different story with \methodname{HGP} \citep{hemati2023understanding} and \methodname{Mixup} \citep{wang2020heterogeneous}, whose rankings are consistent for all \ours instances, %
but drastically change when we average accuracies.
Therefore, average accuracy does lead to significantly different analyses concluding, %
\eg, that \methodname{Mixup} is \emph{worse} than \methodname{SagNet} and \methodname{HGP}, \emph{in disagreement with every other criterion function}.

We can explain this particular case by noting that 
accuracies present \textit{significantly different ranges} across test domains (see \cref{tab:domain-generalization}), %
and hence differences in domains with less variance become less important when computing the average accuracy.
In this case, if we instead normalize the results %
using $\text{norm}_\indexcrit$ (\cref{eq:baseline-methods}), %
we find in the second column of \cref{fig:dom-gen-ranking} that now \methodname{Mixup} significantly outperforms \methodname{HGP} in these domains on average, swapping their relative rankings and better aligning with all \ours instances. %
Similar observation have been made in the context of binary classification, where precision depends on the dataset class ratio and thus needs calibration to compare classifiers across datasets \citep{wissam_master,williams2021effect}.

\subsection{Case 3: Benchmarking}
\label{subsec:case-model-benchmarking}

Finally, we motivate the use of \ours and CDF-normalized \criteria in general in benchmarking settings where, in contrast with the previous use cases, \criteria are not necessarily aggregated into a scalar value, but presented  together to the user in a plot. %
Additional figures can be found in \cref{app:subsec:amlb-results}.

\begin{wrapfigure}[21]{r}{.485\linewidth}
    \centering
    \vspace{-1.\baselineskip}
    \includegraphics{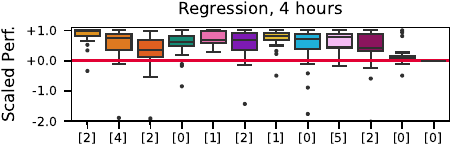}\\
    \includegraphics{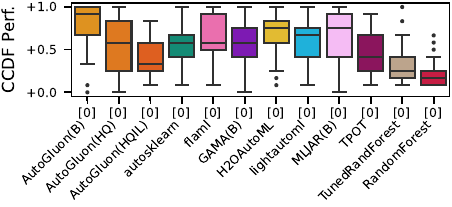}
    \caption{Comparison of AutoML methods on AMLB \citep{amlb} using scaled performance, \methodname{norm}, with a random forest as reference method (red line); %
    	and using a CCDF-transformation \figbottom. Brackets indicate the number of off-view outliers.} \label{fig:amlb-performance}
\end{wrapfigure}
We take the AutoML Benchmark (AMLB) as a nice representative which %
``follows best practices and
avoids common mistakes when comparing %
frameworks'' \citep{amlb}, and
reproduce all figures from the original work, comparing %
\num{15} AutoML methods evaluated on \num{104} \criteria. %
To address incomparability, the authors scaled them %
using $\text{norm}_\indexcrit$ (\cref{eq:baseline-methods}) with a random forest as reference, %
providing then a number of analyses from these \criteria. 
Remarkably, the authors %
encourage the use of CD diagrams and Friedman tests, two methods %
based on relative rankings.

Motivated by the above, a natural step is therefore to use CDF-normalized \criteria. %
\Cref{fig:amlb-performance} shows the same AMLB boxplot using scaled ($\text{norm}_\indexcrit$) %
and CCDF ($1 - \cumulative[\indexcrit](\evarcrit_\indexcrit)$) transformed \criteria. %
We observe that CCDFs produce similar plots and come with several benefits: \itemi~there are no outliers to report, unlike in the original plot; %
\itemii~there is no need to pick an arbitrary reference model; and \itemiii~it provides clear population-based interpretations, \eg, ``on average, \methodname{AutoGluon(B)} \citep{erickson2020autogluon} yields over top-\SI{10}{\percent} performance on the considered \criteria.''
These benefits extend to all AMLB plots, demonstrating that the proposed CDF transformation is a sensible way of normalizing \criteria in general. %

\section{Concluding remarks}

In this work, we have shown the importance of having tools that helps us meaningfully navigate the Pareto front in multi-\criterion ML evaluation, allowing users to perform better-informed decisions, and expanded on which use cases can be interpreted this way. %
To this end, we have highlighted how crucial is to properly normalize all %
\criteria, %
and to have a simple criterion function sensibly mapping DM preferences to the Pareto front. %
We have materialized these insights in the proposed \ours, and extensively demonstrated the potential impact that its adoption can have in areas as fundamental and timely as model selection, comparative model analysis and model benchmarking.

\paragraph{Limitations.} 

It is important to note that the presented results rely on a number of assumptions, and their violation can compromise the performance of \ours, \eg, the continuity of the \criteria or the \iid assumption for the samples.
Also, while the rank-transform works well in our setting, it has been shown not to do so when \eg, modeling variable interactions in statistics \citep{thompson1991note}. The simplicity of \ours might not be appropriate in scenarios where, \eg, two \criteria are perfectly correlated, requiring of DM interaction and, while \ours with $p=\infty$ can reach any solution, in this case the user should be aware and tune $\omegab$ accordingly.
Finally, it is important to stress that relative rankings disregard quantitative changes in \criteria and, \eg, cannot detect phase changes in \criterion values. Therefore, the user should never disregard the original \criteria and thoroughlycheck them before making a final decision.

\paragraph{Future work.}

Our work opens many intriguing venues for future research, \eg, we would be excited to see \ours adapted to active %
settings with humans-in-the-loop, criterion functions that parametrize more complex preferences, a formal systematization of model selection enabled by \ours, or its adoption in public portals such as the Open LLM Leadearboard \citep{open-llm-leaderboard-v2} or the DecodingTrust benchmark~\citep{wang2023decodingtrust}. %
Moreover, future work could expand \ours by using pairwise interactions to normalize the \criteria (rather than marginals), or look at other notions in MOO such as proper Pareto optimality \citep{miettinen1999nonlinear}.

\subsubsection*{Acknowledgments}

We would like to express our gratitude to Christopher K. I. Williams, Peter Flach, N. Siddharth, Nicola Branchini, as well as every member of the APRIL lab for their invaluable feedback while working on this project. 
We also thank Raj Mohan Tumarada whose thesis sparkled the initial ideas for the methodology, and Kaisa Miettinen for helping us understand the role of scaling \criteria in the MOO literature.
This work is part of the project \textit{``Society-Aware Machine Learning: The paradigm shift demanded by society to trust machine learning,''} funded by the European Union and led by IV (\href{https://machinelearning.uni-saarland.de/society-aware-ml/}{ERC-2021-STG}, SAML, 101040177).
AJ received funding from the DFG grant 389792660 as part of \href{https://perspicuous-computing.science}{TRR~248 -- CPEC},
and both AJ and AV were supported by the \textit{``UNREAL: a Unified Reasoning Layer for Trustworthy ML''} project (EP/Y023838/1) selected by the ERC and funded by UKRI EPSR.

 	\vspace{5pt}
    \bibliography{references.clean}
    \bibliographystyle{tmlr}

    \clearpage
    \appendix

    \renewcommand{\partname}{}  %
	\part{Appendix} %
	\parttoc %

    \counterwithin{table}{section}
    \counterwithin{figure}{section}
    \renewcommand{\thetable}{\thesection.\arabic{table}}
    \renewcommand{\thefigure}{\thesection.\arabic{figure}}

\section{Extended related work and existing approaches}
\label{app:sec:baselines}

We review the most relevant methods from different research areas, and discuss their similarities and differences with respect to \ours. 
It is worth stressing that none of the proposed methods, unless otherwise stated, have been applied in the context of multi-\criterion model evaluation in ML.

\paragraph{Normalization functions.}

As mentioned in \cref{sec:problem-statement}, the MOO literature acknowledges that usual optimization objectives are sensitive to the magnitude of their objective values \citep{miettinen1999nonlinear}, and thus they typically addressed it by applying a \emph{component-wise transformation} %
to normalize their objectives. 
Two notorious normalization functions are the following:
\begin{equation} \label{app:eq:baseline-methods}
	\Delta_\indexcrit(\evarcrit_\indexcrit) \coloneqq \frac{\evarcrit_\indexcrit - \ideal{\evarcrit}_\indexcrit}{\ideal{\evarcrit}_\indexcrit}\eqp{,} \quad
	\text{norm}_\indexcrit(\evarcrit_\indexcrit) \coloneqq \frac{\evarcrit_\indexcrit - \ideal{\evarcrit}_\indexcrit}{\nadir{\evarcrit}_\indexcrit - \ideal{\evarcrit}_\indexcrit} \equationPunctuation{,} 
\end{equation}
where $\ideal{\evarcrit}_\indexcrit$ is usually approximated and where $\nadir{\evarcrit}_\indexcrit$ cannot be used for unbounded objectives.
Other normalization functions also exist and, in the context of multi-criteria decision making, ``max normalization'' is also employed (see, \eg, \citep{wang2025multi}). Assuming---as it is usual in their literature---to have a finite set of observations and a \emph{maximization} MOO problem (rather than minimization, as in this work), max normalization is defined as:
\begin{equation}
	\text{max}_\indexcrit(\evarcrit_\indexcrit) = \begin{cases}
		{\evarcrit_\indexcrit}/{\max_\model \evarcrit_\indexcrit}  & \text{if \nth{\indexcrit} \criterion  needs to be maximized,} \\
		{\min_\model \evarcrit_\indexcrit}/{\evarcrit_\indexcrit}  & \text{if \nth{\indexcrit} \criterion  needs to be minimized.}
	\end{cases}
\end{equation}
In our experiments, we noticed that methods using max normalization did not work well. To obtain the results displayed in \cref{fig:llms-figure1}, we adapted the normalization function to a \emph{minimization} MOO problem as:
\begin{equation} \label{app:eq:max-normalization}
	\text{max}_\indexcrit(\evarcrit_\indexcrit) = \begin{cases}
		{\max_\model \evarcrit_\indexcrit}/{\evarcrit_\indexcrit}  & \text{if \nth{\indexcrit} \criterion  needs to be maximized,} \\
		{\evarcrit_\indexcrit}/{\min_\model \evarcrit_\indexcrit}  & \text{if \nth{\indexcrit} \criterion  needs to be minimized.}
	\end{cases}
\end{equation}

\paragraph{Multi-criteria decision making.} 
Continuing with multi-criteria decision making, we compared \ours in \cref{fig:llms-figure1} with the following two baselines described in \citep[Chapter 8]{wang2025multi}. 
First, simple additive weighting (SAW) \citep{saw_original}, which is defined as the weighted average of max-normalized \criteria:
\begin{equation} \label{app:eq:saw}
	\operatorname{SAW}(\omegab) \coloneqq \min_{\indexone \in \seq{\numsamples}} \quad \sum_{\indexcrit=1}^\numobjs \omega_\indexcrit \text{max}_\indexcrit(\evarcrit_\subscript{\indexcrit,\indexone}) \equationPunctuation{.}
\end{equation}
Second, analytic hierarchy process (AHP) \citep{saaty1990make_ahp,saaty1977scaling_ahp} which, intuitively, performs the same max-normalized weighted sum as SAW, but in-between those two steps $\text{max}_\indexcrit(\evarcrit_\indexcrit)$ is substituted by another ``score'' using a spectral criterion. Specifically, AHP is described by the following algorithm:
\begin{enumerate}
	\item Max-normalize each \criterion, $z_\indexcrit \coloneqq \text{max}_\indexcrit(\evarcrit_\indexcrit)$\equationPunctuation{.}
	\item For each \criterion with index $\indexcrit$:
	\begin{enumerate}
		\item Compute a pairwise comparison matrix for the \nth{\indexcrit} \criterion $\mA \in \sR^{\numsamples\times\numsamples}$ as follows:
		\begin{enumerate}
			\item If $z_\subscript{\indexcrit,\indexone} \geq z_\subscript{\indexcrit,\indextwo}$ then $$\emA_\subscript{\indexone,\indextwo} \coloneqq \frac{\ln({z_\subscript{\indexcrit,\indexone}}) - \ln({z_\subscript{\indexcrit,\indextwo}})}{\ln(\max_\indexthree z_\subscript{\indexcrit,\indexthree}) - \ln(\min_\indexthree z_\subscript{\indexcrit,\indexthree})}  \cdot (9 - 1) + 1 \equationPunctuation{.}$$
			\item Otherwise, $\emA_\subscript{\indexone,\indextwo} \coloneqq 1 / \emA_\subscript{\indextwo,\indexone}$\equationPunctuation{.}
		\end{enumerate}
		\item Compute the principal direction of $\mA$, $\vv'_\indexcrit$, such that $\mA\vv'_\indexcrit = \lambda_{\text{max}}\vv'_\indexcrit$\equationPunctuation{.}
		\item Normalize $\vv'_\indexcrit$ to add up to one, $\vv_\indexcrit \coloneqq \vv'_\indexcrit / \sum_\indexone \vv'_\subscript{\indexcrit,\indexone}$\equationPunctuation{.}
	\end{enumerate}
	\item Return the element that minimizes the weighted sum of scores:
	\begin{equation}
		\operatorname{AHP}(\omegab) \coloneqq \min_{\indexone \in \seq{\numsamples}} \quad \sum_{\indexcrit=1}^\numobjs \omega_\indexcrit \vv_\subscript{\indexcrit,\indexone} \equationPunctuation{.}
	\end{equation}
\end{enumerate}
It is worth noting that we have adapted the last step to use a predetermined weight vector $\omegab$. In the original formulation \citep{saaty1990make_ahp,saaty1977scaling_ahp}, the weight vector is derived similar as how the scores $\vv_\indexobj$ were computed, using a given \criterion comparison matrix where the DM makes pairwise comparisons across objectives describing which one is more important to them.

We did not compare with other methods in the multi-criterion decision making literature due to their similarity with other methods. For example, the multiplicative exponent weighting (MEW) method \citep{mew_original} consists on computing the geometric mean (rather than the arithmetic one) of max-normalized \criteria. 
Another example is the ``faire un choix ad\'equat'' (FUCA) method \citep{fernando2011selecting_fuca}, which corresponds to \ours with $p=1$.

\paragraph{Rank-based approaches.}

Following the previous paragraphs, there are other methods than FUCA in different fields that are based on rankings. 
For example, \citet{yamada} proposed in the context of evolutionary algorithms two different utility functions. The first one corresponds to \ours with $p=\infty$ and only using Pareto-optimal points for the rank transformation, as explained in \cref{sec:related-works}. The second one corresponds to a usual weighted $2$-norm using rank-transformed \criteria, which we compare with in \cref{app:fig:compare-pnorms} for $p=8$. 
Similarly, in the context of multi-objective optimization (MOO), \citet{kukkonen2007ranking} proposed to use rank-transformed objectives with $p=1$ (\ie, \ours with $p=1$) or with $p=-\infty$.

\paragraph{Ad-hoc normalization functions.}

Finally, it is worth recalling some of the ways researchers in the past have come up with ways of normalizing \criteria for their specific use cases.
\citet{caruana2004data} proposed in the context of multi-\criteria supervised learning a ``general-purpose metric'' combining squared error, accuracy, and ROC area (therefore denoted as SAR), which is defined as the arithmetic metric of the three quantities. \Cref{sec:problem-statement,sec:experiments} already discussed at length why combining metrics in this way might not be the best idea in general.
In the context of evaluating generative models in tabular data, where each column has a different data type, \citet{nazabal2020hivae} proposed to use the arithmetic mean of the following errors \criteria, depending on the data type:
\begin{enumerate}
	\item Normalized root mean squared error (NRMSE) for numerical variables, defined as:
	\begin{equation}
		\frac{\sqrt{1/\numsamples \sum_\indexone (\ervx_\subscript{\indexcrit,\indexone} - \hat\ervx_\subscript{\indexcrit,\indexone})^2}}{\max_\indextwo \ervx_\subscript{\indexcrit,\indextwo} - \min_\indextwo \ervx_\subscript{\indexcrit,\indextwo} } \equationPunctuation{,}
	\end{equation}
	where $\hat\ervx_\subscript{\indexcrit,\indexone}$ represents the model prediction for the \nth{\indexone} sample and \nth{\indexcrit} column.
	\item Accuracy for the categorical variables.
	\item Displacement for the ordinal variables, defined as the mean absolute error normalized by the range of values that each column can take.
\end{enumerate}
When comparing binary classifiers based on their performance in multiple datasets, it has been shown \citep{wissam_master,williams2021effect} that their precision depends on the ratio of positive and negative samples present in the test set, and therefore it has been proposed to normalize their precision by first re-scaling them, thus enabling the proper comparison of these classifiers across datasets.

In the context of LLM evaluation, it is interesting to revisit the way that the OpenLLM Leaderboard \citep{open-llm-leaderboard-v2}---which we used for our use cases in \cref{sec:intro,sec:experiments}---normalizes the values of their \criteria to perform the score average reported in their website. The following is a summary of the process described in \href{https://huggingface.co/docs/leaderboards/open_llm_leaderboard/normalization}{their website}:
\begin{itemize}
	\item If the task (\ie, criterion) does not subtasks (\eg, GPQA or MMLU-PRO), apply $\text{norm}_\indexcrit$ at the beginning of this section, estimating the nadir point.
	\item If the task has subtasks (\eg, MUSR or BBH), normalize each subtask first, and then average their scores.
	\item Generative tasks require a different approach. The MATH task uses exact match accuracy, and IFEval uses strict accuracy.
\end{itemize}
As a result, the average score reported in the leaderboard (alongside the CO\textsubscript{2} cost) is the average of $\text{norm}_\indexcrit$-normalized \criteria, with some of them being also the average of $\text{norm}_\indexcrit$-normalized sub-\criteria.
Similarly, DecodingTrust \citep{wang2023decodingtrust}, another benchmark for LLM evaluation, normalizes each of its eight \criteria in a different way to make them comparable, as we describe in detail in \cref{app:subsec:decodingtrust}.

\section{Experimental details and additional results}
\label{app:sec:exp-details}

In this section, we provide all details to reproduce the experiments presented in the manuscript, as well as additional results which were omitted from the main paper due to space constraints.

\subsection{Synthetic evaluation}

\begin{wrapfigure}[18]{R}{.5\linewidth}
    \centering
    \vspace{-1.5\baselineskip}
    \includegraphics{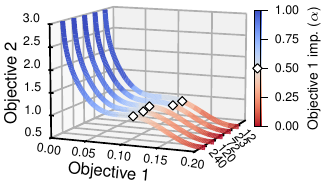} 
    \caption{
        Synthetic Pareto front showing the Pareto front using \ours with $p=\infty$ as we change the number of sampled points. While it can be observed a deterioration on the estimated Pareto front (see quantized colors as we reduce \numsamples), \ours offers a robust estimator even with 12 datapoints.}
    \label{app:fig:3d-synthetic}
\end{wrapfigure}

As we describe in the main text, for the synthetic experiment we consider the following parametric curve:
 \begin{equation}
    \evarcrit_2 = 0.25 \cos(39\evarcrit_1^{0.85}) - \log(\evarcrit_1) - 0.46 \,,
\end{equation}
where $\evarcrit_1 \sim \uniform(0.02, 0.2)$\,. 
As a result, we end up with a non-convex Pareto front with a flat area around $\evarcrit_1 = 0.1$, and two \criteria with significantly different distributions.
Moreover, the distribution of both objectives are significantly different. 
Specifically, the first objective is uniformly distributed, while the second one is precisely the plotted curve (if we flipped it to have the second objective as the x-axis), therefore being heavy tailed with most density lying in the $\interval{0}{0.2}$ interval.
The uneven and long-stretch of the domain of the second objective explains why, despite applying $\text{norm}_\indexcrit$, we still get a biased optimization problem in \cref{fig:3d-problem-statement}, as discussed in \cref{sec:problem-statement}.

\subsubsection{Additional results}

\paragraph{How robust are we to sample size?}

Despite having a closed-form expression for the variance of our estimator $\evarcdf_\indexobj$ in \cref{subsec:finite-sample-approx}, we empirically show in \cref{app:fig:3d-synthetic} the estimated Pareto front using \ours with $p=\infty$ as a function of the first-objective importance, $\alpha$, as we change the total number of points sampled to estimate it, \numsamples. 
We can observe that, despite considerably reducing the number of samples from \num{240} to \num{12} datapoints, the estimate given by \ours remains perfectly consistent.

\subsection{Open LLM Leaderboard: Navigating the LLM performance-cost Pareto front}
\label{app:subsec:llms-details}

\paragraph{Dataset details.}

In order to conduct our experiments, we retrieved the publicly available results from the Open LLM Leaderboard \citep{open-llm-leaderboard-v2} using Huggingface's dataset Python package and, for reproducibility purposes, saved a local copy with the state as of the \nth{9} of January 2025.
From the \num{2929} total LLMs, we discard those which were not publicly available on Huggingface's hub. %
This leave us with a total of \num{2148} models, which we use to conduct the experiments described in this work.

\paragraph{Experimental details.}

As explained in the main text, we consider all reported values as \criteria. Namely, we take as \criteria the CO\textsubscript{2} emissions and all \num{6} benchmark performance scores computed on the following datasets: IFEval \citep{zhou2023instructionfollowingevaluationlargelanguage}, BBH \citep{suzgun2022challengingbigbenchtaskschainofthought}, MATH \citep{hendrycks2021measuringmathematicalproblemsolving}, GPQA \citep{rein2023gpqagraduatelevelgoogleproofqa}, MuSR \citep{sprague2024musrtestinglimitschainofthought}, and MMLU-Pro \citep{wang2024mmluprorobustchallengingmultitask}.
Then, we use \ours with $p=\infty$ to produce both \cref{fig:llms-figure1,fig:llms-leaderboard}, setting the values of $\omegab$ according to the importance given to CO\textsubscript{2} emissions, $\alpha$, as $\omegab \coloneqq \irow{\alpha, \frac{1-\alpha}{6}, \dots, \frac{1-\alpha}{6}}$\,.
To create these figures, we take \num{10000} values of $\alpha$ evenly-spaced in the unit interval and, since different values of $\alpha$ can provide us with the same model, use their range-average (\cref{fig:llms-figure1}) or maximum (\cref{fig:llms-leaderboard}) as the value to colour the selected LLMs in the figures.
There are two more details worth-discussing.
First, in \cref{fig:llms-figure1} we use \ours over two \criteria (the average score and CO\textsubscript{2} emissions) just so that the models selected by all criterion functions lied exactly in the plotted Pareto front, since Pareto-optimal models selected with all $\numcrit = 7$ \criteria may not be Pareto-optimal when considering this bidimensional representation. 
Second, we use as y-axis for \cref{fig:llms-leaderboard} the CDF of the $p$-norm computed using the CDF-transformed performance criteria (\ie, of the vector used with \ours, excluding the CO\textsubscript{2} dimension), since this represents much more closely the CDF-space that \ours navigates.

\subsubsection{Additional results} \label{app:subsec:llm-additionalresults}

\paragraph{Retrieved models.} 
Complementing \cref{fig:llms-leaderboard}, we present here the quantitative results of those LLMs selected with \ours. In the table we report the reported benchmark scores, a summary of their benchmark performance and CO\textsubscript{2}, the CDF values found by \ours (same as in \cref{fig:llms-leaderboard}), and the value of $\alpha$ used to select these models. 
As it can be observed, \ours allows us to meaningfully navigate the performance-cost trade-off in the LLM space.
Answering the initial question we posed in \cref{sec:intro}, if we were a practitioner trying to select a balanced LLM in terms of its performance and cost without further prior expectations, \emph{we} would proceed in this case by using \ours with $p=\infty$ and $\alpha=0.5$, which would yield us a model, \href{https://huggingface.co/unsloth/Phi-3-mini-4k-instruct}{unsloth/Phi-3-mini-4k-instruct}, in the top-\SI{9}{\percent} of LLMs in terms of benchmark performance, and top-\SI{8}{\percent} in terms of CO\textsubscript{2} emissions.

{
    
    \sisetup{table-format=2.2, round-mode = places, round-precision = 2, detect-all}
    \begin{table*}[h]
        \centering
        \caption{Quantitative results of the LLMs highlighted in \cref{fig:llms-leaderboard} from the Open LLM Leadearboard \citep{open-llm-leaderboard-v2} using \ours with $p=\infty$, as we change the importance of CO\textsubscript{2} consumption. Rather than using the average, the CDF value for the performance computes the weighted $\infty$-norm of the CDF-transformed benchmark results (\ie, the value used with \ours but separating CO\textsubscript{2} from the rest of \criteria).}  \label{tab:llms-selected-full}
        \resizebox{\linewidth}{!}
        {
            \begin{tabular}{lSSSSSSSSSSS}
                \toprule
                & \multicolumn{6}{c}{Benchmarks scores} & \multicolumn{2}{c}{Summary} &  \multicolumn{2}{c}{CDF values} & \\ \cmidrule(lr){2-7} \cmidrule(lr){8-9} \cmidrule(lr){10-11}
                 & {IFEval} & {BBH} & {MATH} & {GPQA} & {MUSR} & {MMLU-PRO}  & {Average} & {CO\textsubscript{2} cost} & {\multirow{2}{*}{\shortstack{Perf.\\\small($p=\infty$)}}} & {\multirow{2}{*}{\shortstack{CO\textsubscript{2}\\cost}}} & \\
                Full model name & {(\%)} & {(\%)} & {(\%)} & {(\%)} & {(\%)} & {(\%)} & {(\%)} & {(kg)} &  &  & {$\alpha$} \\
                \midrule
                \href{https://huggingface.co/dfurman/CalmeRys-78B-Orpo-v0.1}{dfurman/CalmeRys-78B-Orpo-v0.1} & 81.632734 & 61.924764 & 40.709970 & 20.022371 & 36.372135 & 66.801492 & 51.243911 & 12.996767 & 0.000000 & 0.952492 & 0.005005 \\
                \href{https://huggingface.co/maldv/Qwentile2.5-32B-Instruct}{maldv/Qwentile2.5-32B-Instruct} & 73.931613 & 57.205878 & 38.066465 & 17.897092 & 19.961979 & 54.214687 & 43.546286 & 3.532542 & 0.006986 & 0.868188 & 0.021021 \\
                \href{https://huggingface.co/sometimesanotion/Qwen2.5-14B-Vimarckoso-v3}{sometimesanotion/Qwen2.5-14B-Vimarckoso-v3} & 72.565238 & 48.581587 & 34.441088 & 17.337808 & 19.385938 & 48.258348 & 40.095001 & 1.928625 & 0.013507 & 0.785748 & 0.029029 \\
                \href{https://huggingface.co/hotmailuser/FalconSlerp3-7B}{hotmailuser/FalconSlerp3-7B} & 60.962358 & 36.834016 & 27.416918 & 9.172260 & 15.901302 & 34.748079 & 30.839155 & 0.606662 & 0.046111 & 0.192361 & 0.211211 \\
                \href{https://huggingface.co/unsloth/Phi-3-mini-4k-instruct}{unsloth/Phi-3-mini-4k-instruct} & 54.402762 & 36.732473 & 15.407855 & 9.731544 & 13.118750 & 33.676862 & 27.178374 & 0.469533 & 0.079180 & 0.089893 & 0.499499 \\
                \href{https://huggingface.co/icefog72/Ice0.37-18.11-RP}{icefog72/Ice0.37-18.11-RP} & 49.721628 & 31.042850 & 6.419940 & 8.277405 & 12.207552 & 23.814273 & 21.913941 & 0.414513 & 0.213321 & 0.067070 & 0.658659 \\
                \href{https://huggingface.co/h2oai/h2o-danube3.1-4b-chat}{h2oai/h2o-danube3.1-4b-chat} & 50.211217 & 10.942063 & 2.114804 & 4.697987 & 10.202865 & 19.095375 & 16.210718 & 0.299141 & 0.595715 & 0.032604 & 0.821822 \\
                \href{https://huggingface.co/postbot/gpt2-medium-emailgen}{postbot/gpt2-medium-emailgen} & 14.920300 & 3.673700 & 0.000000 & 1.342282 & 6.889323 & 1.632683 & 4.743048 & 0.078186 & 0.863065 & 0.004192 & 0.973974 \\
                \bottomrule
            \end{tabular}
        }
    \end{table*}
    
}

\paragraph{Differences in $p$-norms.}

\begin{figure}[h]
	\centering
	\begin{subfigure}[t]{.49\linewidth}
		\includegraphics{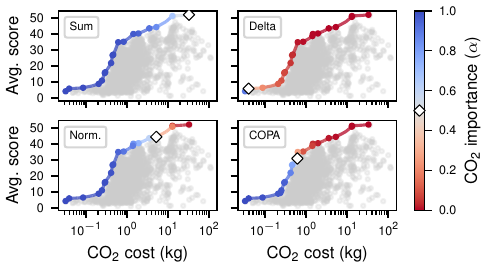}
		\caption{Proposed norm: $\left(\, \sum_{\indexcrit=1}^\numcrit \abs{\omega_\indexcrit \phi_\indexcrit(\evarobj_\indexcrit)}^p \,\right)^{1/p}$.}
	\end{subfigure}%
	\hfill%
	\begin{subfigure}[t]{.49\linewidth}
		\includegraphics{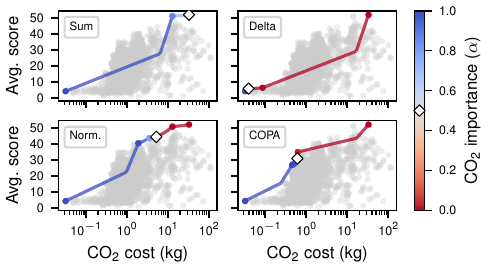}
		\caption{Usual weighted $p$-norm: $\left(\, \sum_{\indexcrit=1}^\numcrit \omega_\indexcrit \abs{\phi_\indexcrit(\evarobj_\indexcrit)}^p \,\right)^{1/p}$.}
	\end{subfigure}
	\caption{LLM experiment comparing the proposed and the usual weighted $p$-norms, with $p=8$ and taking \num{10000} evenly-spaced values for $\alpha$. As discussed after introducing \cref{eq:weighted-pnorms}, the usual weighted $p$-norm is not well-suited for our purposes, as CDF-transformed \criterion lie in the unit interval and quickly vanish.}
    \label{app:fig:compare-pnorms}
\end{figure}

To show the differences between using as criterion function the usual $p$-norm or the one proposed in this paper (\cref{eq:weighted-pnorms}), \cref{app:fig:compare-pnorms} shows the same figure as in \cref{fig:llms-figure1}, comparing the proposed norm in \cref{eq:weighted-pnorms} and the usual weighted $p$-norm with four different normalization functions $\phib$. 
Note that we do not use $p=\infty$ as in the original figure since, for the usual weighted $p$-norm, $\norm{\varcdf}_{\infty,\omegab} = \norm{\varcdf}_\infty$, while for the proposed norm it corresponds to the weighted Tchebycheff problem, as discussed in \cref{subsec:p-norms}.

\paragraph{Using piece-wise criterion functions.}

\begin{figure}[h]
	\centering
	\begin{subfigure}[t]{.49\linewidth}
		\includegraphics{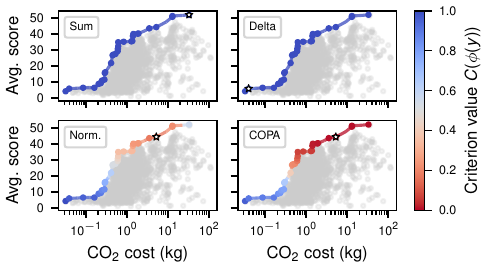}
	\end{subfigure}%
	\hfill%
	\begin{subfigure}[t]{.49\linewidth}
		\includegraphics{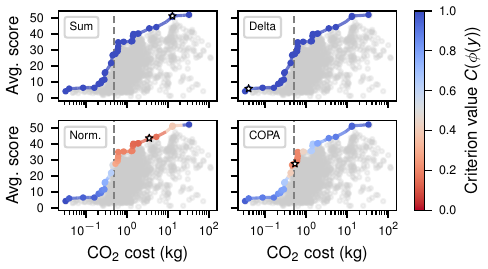}
	\end{subfigure}\\
	\begin{subfigure}[c]{.49\linewidth}
		\caption{$\utility(\varobj) = \norm{\phib(\varobj)}_\subscript{8,0.01}$}
	\end{subfigure}%
	\hfill%
	\begin{subfigure}[c]{.49\linewidth}
		\caption{$\utility(\varobj) = \begin{cases}\norm{\phib(\varobj)}_\subscript{8,0.01} & \text{if } \evarobj_2 < \SI{0.5}{\kilo\gram} \\ \norm{\phib(\varobj)}_\subscript{8,0.99} & \text{if } \evarobj_2 \geq \SI{0.5}{\kilo\gram} \end{cases}$}
	\end{subfigure}
	\caption{LLM experiment where we compare the model retrieved using the proposed weighted $p$-norm, and a piece-wise criterion function that depends on the \criteria values. Colors represent the norm value of each model in the Pareto front (\ie, the value of the criterion function), and we look for the point with minimum norm (marked with a star).The importance of having comparable \criteria is clear, as only \ours finds a point near the decision boundary despite the highly skewed weights in both norms.}
    \label{app:fig:piecewise-function}
\end{figure}

In the main pages, we exclusively consider criterion functions as weighted norms, as proposed in \cref{eq:weighted-pnorms}. 
To showcase that this is not a real restriction---in fact, we can use any sensible criterion function which we can evaluate---we show in \cref{app:fig:piecewise-function} the same experiment as in \cref{app:fig:compare-pnorms} where we replaced the criterion function such that it depends on the CO\textsubscript{2} footprint of the model we are evaluating.
We take this experiment to the extreme, such that the criterion function gives almost-zero importance to CO\textsubscript{2} cost if the model consumes less than \SI{0.5}{\kilo\gram}, and almost no-importance to performance otherwise. As we can see, it is still crucial to have semantically comparable \criteria, as the rest of normalization functions find a solution far from the decision boundary despite the highly-skewed weights. Furthermore, this experiment shows that we can use any criterion function sensible for the DM.

\subsection{Navigating the fairness-accuracy trade-off}

\paragraph{Experimental details.}

We reproduce the CelebA \citep{liu2015celeba} experiment from \citep{maheshwari2022fairgrad} using their proposed FairGrad algorithm, which code is publicly available at \href{https://github.com/saist1993/fairgrad/blob/main/examples/simple_classification_dataset.py}{github.com/saist1993/fairgrad}, and run this experiment with \num{10} random initializations and \num{24} different values of $\epsilon$ (the hyperparameter of FairGrad that represents the desired fairness upper-bound), namely: %
\begin{align*}
     \epsilon \in \{ & 0.001, 0.002, 0.003, 0.004, 0.005, 0.006, 0.007, 0.008, 0.009, \\ 
     & 0.01 , 0.02 , 0.03 , 0.04 , 0.05 , 0.06 , 0.07 , 0.08 , 0.09 , \\ 
     & 0.   , 0.1  , 0.2  , 0.3  , 0.5  , 1.\}
\end{align*}
This leave us with a total of \num{240} models.
To produce \cref{fig:fair-ml}, we use \ours with $p=\infty$ and \num{50} values of $\alpha$ evenly-spaced in the unit interval. For the constrained case, we simply drop those points that do not match the requirements for accuracy (being larger than \num{0.845}) and fairness (having an equal opportunity value smaller than \num{0.02}) before selecting any models with \ours.
Of course, to compute the rankings of the accuracy, we take into account that it needs to be maximized and used the opposite order relation. Similarly, when we applied other normalization functions (see below), we employ the error rate (rather than the accuracy), so that it has to be minimized.

\subsubsection{Additional results}

\begin{figure}[t]
    \centering
    \begin{subfigure}{.45\linewidth}
        \centering
        \includegraphics{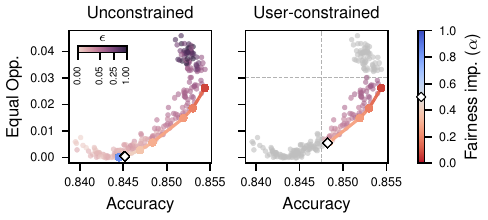}
        \caption{Naively adding up all \criteria.}
    \end{subfigure}%
    \hspace{.025\linewidth}
    \begin{subfigure}{.45\linewidth}
        \centering
        \includegraphics{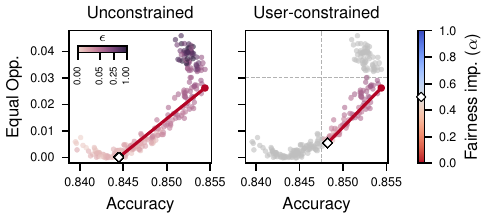}
        \caption{Adding up \criteria with $\phi_\indexcrit = \Delta_\indexcrit$.}
    \end{subfigure} \\[\baselineskip]
    \begin{subfigure}{.45\linewidth}
        \centering
        \includegraphics{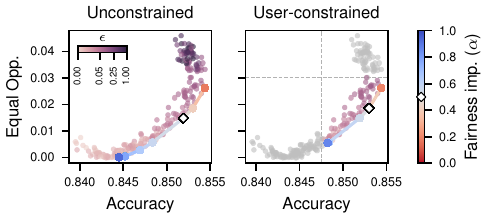}
        \caption{Adding up \criteria with $\phi_\indexcrit = \text{norm}_\indexcrit$.}
    \end{subfigure}%
    \hspace{.025\linewidth}
    \begin{subfigure}{.45\linewidth}
        \centering
        \includegraphics{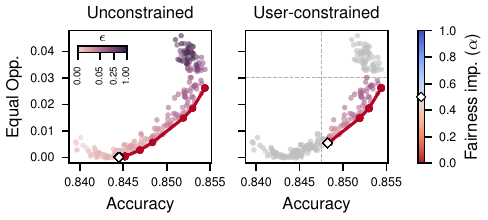}
        \caption{Using SAW \citep{saw_original}.}
    \end{subfigure}\\[\baselineskip]
    \begin{subfigure}{.45\linewidth}
    \centering
    \includegraphics{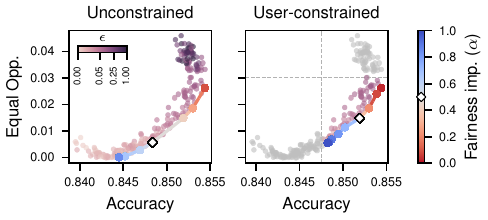}
    \caption{Using AHP \citep{saaty1990make_ahp,saaty1977scaling_ahp}.}
    \end{subfigure}%
    \hspace{.025\linewidth}
    \begin{subfigure}{.45\linewidth}
    \centering
    \includegraphics{figs/fairness-case-cdf.pdf}
    \caption{Using \ours with $p=\infty$.}
    \end{subfigure}%
    \caption{We reproduce the fair ML experiment from \cref{subsec:case-model-selection} using different baselines. We can observe that \ours meaningfully navigates the Pareto front, with all other approaches being biased towards one of the extreme solutions to some extent. Indeed, $\Delta_\indexcrit$ only reaches the two extreme solution despite sampling \num{50} evenly-spaced values for $\alpha$.}
    \label{app:fig:all-fairness}
\end{figure}

We show in \cref{app:fig:all-fairness} the same plot as in \cref{fig:fair-ml}, using all the considered normalization functions and baselines.
Similarly to what we observed in the introductory example in \cref{fig:llms-figure1}, all other methods are biased towards minimizing one of the \criteria. It is worth noting, however, than in this case AHP \citep{saaty1990make_ahp,saaty1977scaling_ahp} and norm do a good job.

\subsection{Comparative model analysis experiments}
\label{app:subsec:model-analysis-results}

\begin{figure}[t]
	\centering
	\begin{subfigure}{.5\linewidth}
		\centering
		\includegraphics{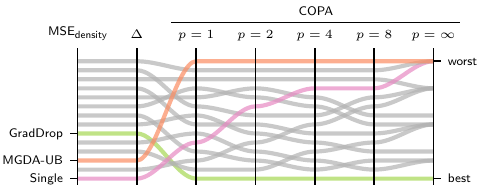}
		\caption{MTL experiment.}
	\end{subfigure}%
	\begin{subfigure}{.5\linewidth}
		\centering
		\includegraphics{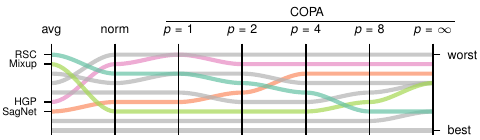}
		\caption{Domain generalization experiment.}
	\end{subfigure}%
	\caption{Reproductions of \cref{fig:mtl-ranking,fig:dom-gen-ranking} where we do not untie methods. As it can be observed, conclusions drawn in the main paper do not change and untying only serves aesthetic purposes.}
	\label{app:fig:no-untie}
\end{figure}

\paragraph{Experimental details.} For the figures shown in \cref{subsec:case-model-analysis}, we retrieved the results reported by two selected works. 
In particular, we took the values reported in the second half of Table 5 from the work of \citet{javaloy2022rotograd} for the MTL experiment, and values reported in Table 4 of \citet{hemati2023understanding} for the domain generalization experiment of the main text. 
From these values, we simply re-rank them using the different criterion functions discussed in the main paper, and highlight those which we consider are interesting for the discussion we carry out in the main manuscript.
To ease visualization, as ties are more frequent in $p=1$ and $p=\infty$---especially when we have only a handful of models---we untie by using the ranking of $p=8$ as a secondary criterion. That is, if two models tie, we rank those by their performance with $p=8$. We plot the figures without untying in \cref{app:fig:no-untie} for the sake of transparency, showing that conclusions do not change.
We use equal weights for all versions of \ours.
One important detail is that, for the domain generalization case, we kept only the top methods, as the rest do not add anything more to the discussion and make the plot more difficult to read.

\subsubsection{Additional results}

{
    
    \sisetup{table-format=2.2, round-mode = places, round-precision = 2, detect-all}
    \begin{table}[t]
        \centering
        \caption{{Different effective ranges explain the differences in rankings of the domain generalization experiment}. The table shows the effective range of each domain accuracy, and the performance of \methodname{Mixup} and \methodname{HGP} for the raw and normalized ($\text{norm}_\indexcrit$, \cref{eq:baseline-methods}) domain accuracies, respectively.} 
        \label{tab:domain-generalization}
        {
            \begin{tabular}{c@{\hspace{\tabcolsep}}lSSSSS} 
                \toprule 
                & & \text{VLC} & \text{PACS} & \text{OfficeHome} & \text{DomainNet} & \text{Avg} \\ \midrule
                \multicolumn{2}{l}{Min. acc.} & 76.3 & 78.8 & 60.2 & 23.4 & {\minusmark} \\
                \multicolumn{2}{l}{Max. acc.} & 79.3 & 84.8 & 68.5 & 41.4 & {\minusmark} \\ \midrule
                \multirow{2}{*}{\rotatebox[origin=c]{90}{\shortstack{\small Acc.}}} %
                & Mixup & 77.7 & 83.2 & 67 & 38.5 & 66.6 \\
                & HGP & 76.7 & 82.2 & 67.5 & 41.1 & 66.875  \\ 
                \midrule
                \multirow{2}{*}{\rotatebox[origin=c]{90}{\shortstack{\small Norm.}}} %
                & Mixup & 46.66666667& 73.33333333 & 81.92771084 & 83.88888889 & 71.45414993 \\
                & HGP & 13.33333333 & 56.66666667 & 87.95180723 & 98.33333333 & 64.07128514 \\
                \bottomrule
            \end{tabular}
        }
    \end{table}    
}

\begin{figure}[t]
    \centering
    \begin{subfigure}{.45\linewidth}
        \centering
        \includegraphics{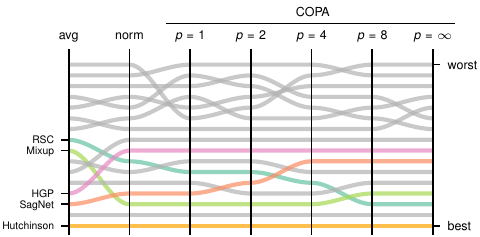}
        \caption{Table 4 from \citet{hemati2023understanding}.}
        \label{app:subfig:dom-gen-table4}
    \end{subfigure}%
    \hspace{0.025\linewidth}
    \begin{subfigure}{.45\linewidth}
        \centering
        \includegraphics{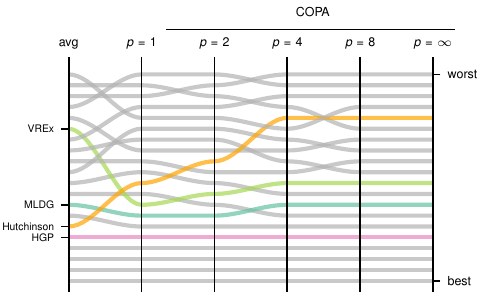}
        \caption{Table 9 from \citet{hemati2023understanding}.}
        \label{app:subfig:dom-gen-table9}
    \end{subfigure}%
    \caption{Ranking of the domain generalization methods considered by \citet{hemati2023understanding} as we use different criterion functions to rank them. We can appreciate a significant change of rankings, and the average accuracy in particular being highly inconsistent with all versions of \ours. We highlight those methods used for the discussion in the text.}
\end{figure}

As mentioned just above, we discarded some methods in the domain generalization figure of the main text (\ie, \cref{fig:dom-gen-ranking}). 
For completeness, we show in \cref{app:subfig:dom-gen-table4} the full figure with all methods included, and highlighting \methodname{Hutchinson}, the second method proposed by the authors, along \methodname{HGP}.
Also, we show in \cref{app:subfig:dom-gen-table9} the same figure but using as data the one reported in Table 9 from \citet{hemati2023understanding} (instead of Table 4). 
This table was reported in the supplementary material, and the difference between both tables is the method used to select hyperparameters, with all methods but those proposed by this particular work (\ie, \methodname{HGP} and \methodname{Hutchinson}) improving their performance.
More crucially, we show once again the huge discrepancies in ranking between using the average accuracy and any of the \ours versions. This time, we also report \methodname{Hutchinson}, which is the best method for all criterion functions in \cref{app:subfig:dom-gen-table4}, and the fourth to worst method in \cref{app:subfig:dom-gen-table9}. 
We can again observe how much our final conclusions can change in \cref{app:subfig:dom-gen-table9}, where the fourth to worst method in terms of average domain accuracy, \methodname{VREx} \citep{krueger2021out}, is better than \methodname{Hutchinson} in all instances of \ours.
To finalize, we consider important to report that, in both figures, the first gray line (\ie, the second-best and best methods, respectively) correspond to the domain generalization method named \methodname{CORAL} \citep{sun2016deep}.

\subsection{AutoML Benchmarking (AMLB) experiment}
\label{app:subsec:amlb-results}

\paragraph{Experimental details.} 

To demonstrate the out-of-the-box utility of \ours and its two components, we reproduce some of the plots from the AutoML Benchmark from \citet{amlb}. 
To achieve this, we simply modify the Jupyter notebook publicly available at \href{https://github.com/PGijsbers/amlb-results/blob/main/notebooks/visualization.ipynb}{github.com/PGijsbers/amlb-results}, and add a few lines of code to compute \ours as proposed in this work.

\subsubsection{Additional results}

To complement \cref{fig:amlb-performance} from the main text, we provide here side-by-side comparisons of more figures reported by \citet{amlb}, further reinforcing the argument of broadly adopting CDF-transformed \criteria for general cases.

\begin{figure}[t]
    \centering
    \begin{subfigure}{.45\linewidth}
        \centering
        \includegraphics{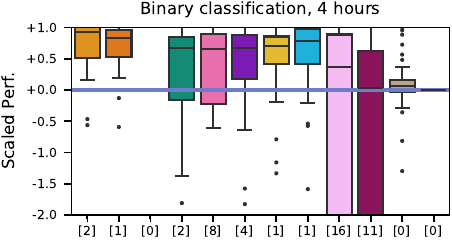}\\%
        \includegraphics{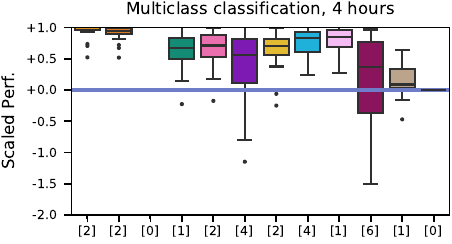}\\%
        \includegraphics{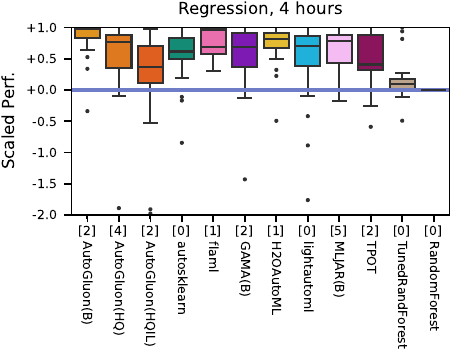}
        \caption{Using scaled performance.}
    \end{subfigure}%
    \hspace{0.025\linewidth}
    \begin{subfigure}{.45\linewidth}
        \centering
        \includegraphics{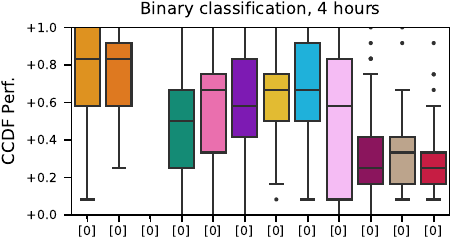}\\%
        \includegraphics{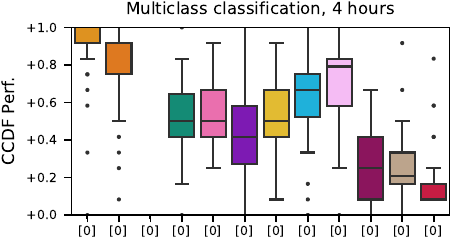}\\%
        \includegraphics{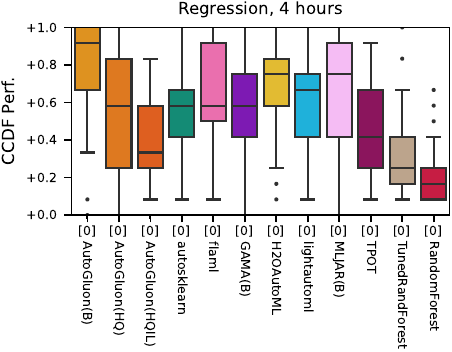}
        \caption{Using CCDF values.}
    \end{subfigure}%
    \caption{We reproduce Fig.~3 from \citet{amlb} in \captiona using their proposed scaled performance, and we show the same figure in \captionb but using complementary CDF values (CCDF, one minus the CDF value). The same advantages as those discussed in \cref{subsec:case-model-benchmarking} can be observed here.}
    \label{app:fig:amlb-boxplots}
\end{figure}

\begin{figure}[t]
    \centering
    \begin{subfigure}{.48\linewidth}
        \centering
        \includegraphics{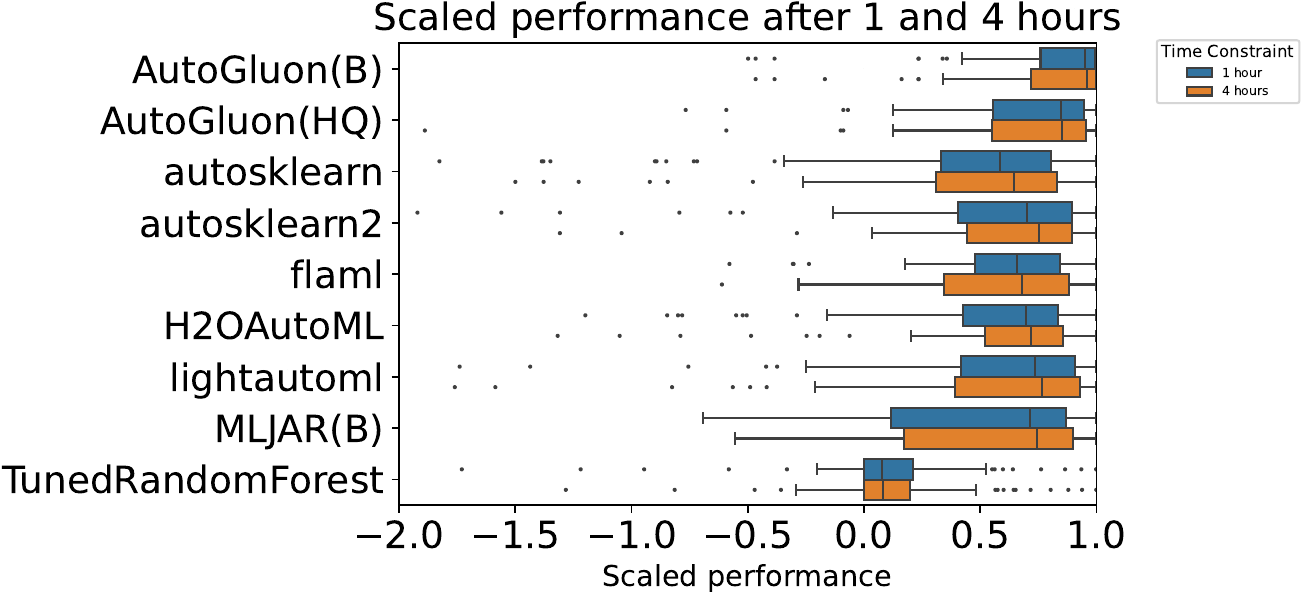}%
        \caption{Using scaled performance.}
    \end{subfigure}%
    \hspace{0.025\linewidth}
    \begin{subfigure}{.48\linewidth}
        \centering
        \includegraphics{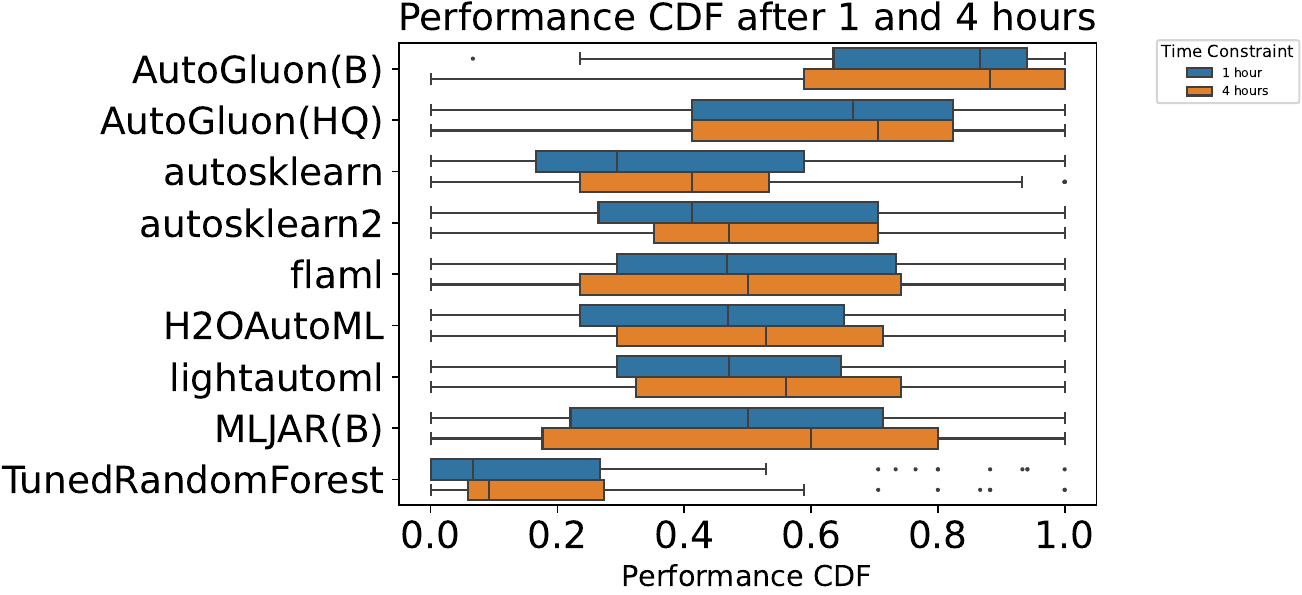}%
        \caption{Using CCDF performance.}
    \end{subfigure}%
    \caption{We reproduce Fig.~4 from \citet{amlb} in \captiona using their proposed scaled performance, and we show the same figure in \captionb but using complementary CDF values (CCDF, one minus the CDF value). The same advantages as those discussed in \cref{subsec:case-model-benchmarking} can be observed here.}
    \label{app:fig:amlb-single-boxplot}
\end{figure}

In particular, we show in \cref{app:fig:amlb-boxplots} the same three figures as Figure 3 from the original work, where the same advantages when using the proposed CDF transformation, as those discussed in the main text (see \cref{subsec:case-model-benchmarking}), can be observed here.
Furthermore, we show in \cref{app:fig:amlb-single-boxplot} Figure 4 from the original work, where all \num{104} \criteria are used, further showcasing the benefits of the proposed transformation.

\begin{figure}[t]
    \centering
    \begin{subfigure}{.45\linewidth}
        \centering
        \includegraphics{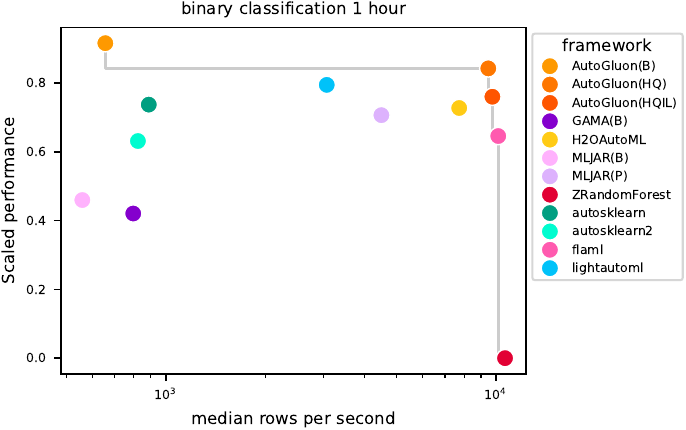}\\%
        \includegraphics{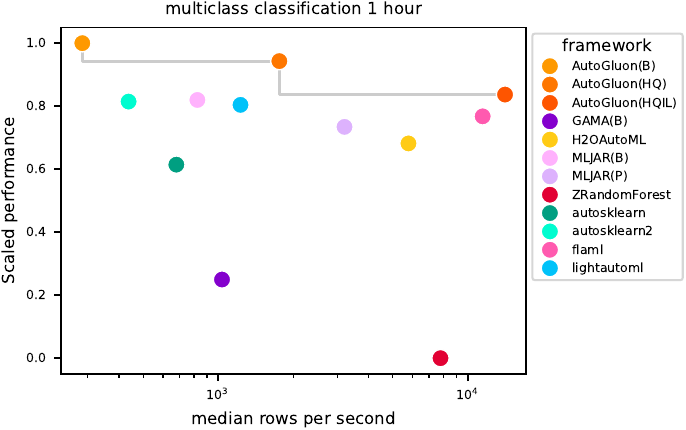}\\%
        \includegraphics{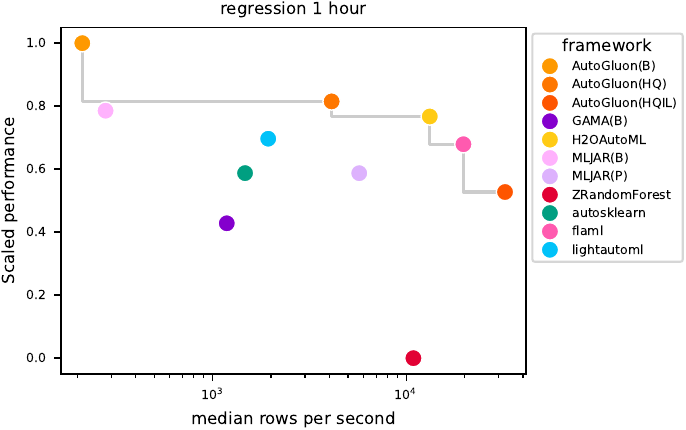}
        \caption{Using scaled performance.}
    \end{subfigure}%
    \hspace{0.025\linewidth}
    \begin{subfigure}{.45\linewidth}
        \centering
        \includegraphics{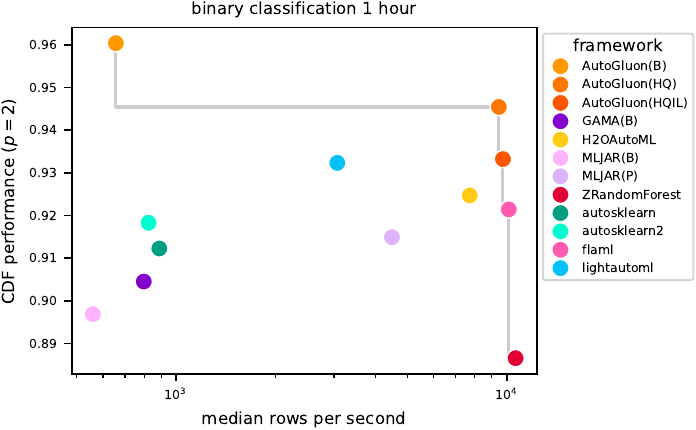}\\%
        \includegraphics{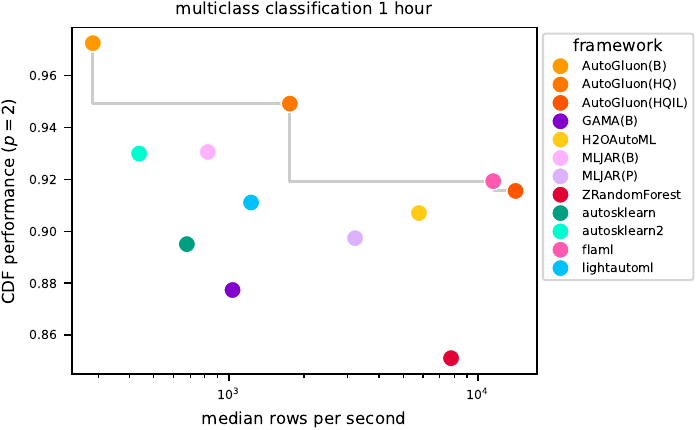}\\%
        \includegraphics{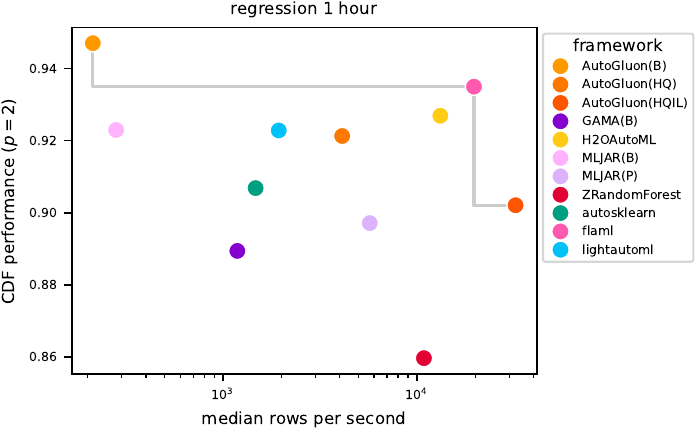}
        \caption{Using CCDF values.}
    \end{subfigure}%
    \caption{We reproduce Fig.~7 from \citet{amlb} in \captiona using their proposed scaled performance, and we show the same figure in \captionb but using complementary CDF values (CCDF, one minus the CDF value).}
    \label{app:fig:amlb-pareto-fronts}
\end{figure}

Finally, we also reproduce Figure 7 from the original publication in \cref{app:fig:amlb-pareto-fronts}, where different Pareto plots are generated according to the type of tasks, showing the performance-speed trade-off, similar in spirit to \cref{fig:llms-figure1} in this work.
Here, we use \ours with $p=2$ and equal weights.
We can observe that, while some of the figures are quite similar, \eg, binary classification in the top row, some others differ significantly, \eg, regression in the bottom row, where \ours reports two less Pareto-optimal models.
Beyond the differences in using scaled \versus CDF-transformed \criteria, which we have extensively discussed during this paper, and showed the significant advantages of employing the latter, the differences in the number of Pareto-optimal models is due to the fact that the Pareto front is computed \emph{after} aggregating the performance metrics.
This is in stark contrast with the approach taken in this work (except for \cref{fig:llms-figure1} for visualization purposes, see \cref{app:subsec:llms-details}), where we compute Pareto-optimal points on the space of all \criteria.
As we have been arguing during this work, \ours allows us to meaningfully navigate the Pareto front, enabling the creation of plots such as those reported in this work (\eg, \cref{fig:llms-figure1,fig:llms-leaderboard,fig:fair-ml}), which are significantly more informative than those reported before our work, as it can be clearly observed in \cref{app:fig:amlb-pareto-fronts}.

\FloatBarrier
\subsection{DecodingTrust: Navigating the LLM trustworthiness Pareto front} \label{app:subsec:decodingtrust}

\paragraph{Dataset details.} 

We look once more into the LLM space and, this time, we focus on the DecodingTrust leaderboard~\citep{wang2023decodingtrust}. In contrast with the Open LLM leaderboard~\citep{open-llm-leaderboard-v2}, DecodingTrust focuses in assessing the trustworthiness of LLM models, rather than on sheer performance. To this end, the authors design a comprehensive list of prompts that the model should successfully answer to. For the interest of this work, it suffices to say that DecodingTrust evaluates LLMs on \num{8} completely different aspects and the authors had to come up with different ad hoc normalization functions for each of the \criteria so that they all lie in the $\interval{0}{100}$ interval. 
Below there is a summary of the formulas employed by the authors and, while we do not give context for the variables shown in the equations, we want to stress the diversity of scores and normalizations that the authors had to propose to make \criteria more comparable. For further details on the normalization functions and, more in general, DecodingTrust, refer to \citep[App. I]{wang2023decodingtrust}:

\begin{itemize}
	\item \emph{Toxicity}: $\displaystyle 1 - \frac{1}{2\sum_i \abs{D_i}} \sum_{i=1}^4 \sum_{x \in D_i} \left( f(x^*_\text{adv} ; x) + f(x^*_\text{benign}; x) \right)\equationPunctuation{.}$
	\item \emph{Stereotype bias}: $$\frac{S_\text{benign} + S_\text{untargeted} + S_\text{targeted}}{3} \quad\text{where}\quad S_\text{scenario} = 1 - (\sum_{i=1}^{n_\text{ST}} \sum_{j=1}^{n_\text{DG}} S_{ij}) / (n_\text{ST} n_\text{DG})\equationPunctuation{.}$$
	\item \emph{Adversarial robustness}: $\displaystyle \frac{\sum_{i=1}^T \text{acc}_i * d_i}{\sum_{i=1}^T d_i} \equationPunctuation{.}$
	\item \emph{Out-of-distribution robustness}: $(\text{ACC}_\text{style} + \text{Reliability}_\text{OOD} + \text{ACC}^\text{icl}_\text{style} + \text{ACC}^\text{icl}_\text{domain}) / 4$ where:
	\begin{align*}
		& \text{ACC}_\text{style} = \frac{1}{S} \sum_{s=1}^S \text{acc}_s \equationPunctuation{,} \\
		&\text{Reliability}_\text{OOD} = \frac{\text{Reliability}_\text{2023} + \text{Reliability}_\text{2023idk}}{2} \\
		&\text{Reliability}_\text{setting} = \text{RR}_\text{setting} + (1 - \text{RR}_\text{setting}) * \text{macc}_\text{setting} \equationPunctuation{,} \\
		& \text{ACC}_\text{setting}^\text{icl} = \frac{1}{D * N} \sum_{d=1}^D \sum_{i=1}^N \text{acc}_{di}^\text{setting} \equationPunctuation{.}
	\end{align*}
	\item \emph{Robustness to adversarial demonstrations}: $ (s^\text{(cf)} + s^\text{(sc)} + s^\text{(bkd)}) / 3$ where:
	\begin{align*}
		& s^\text{(cf)} = \frac{1}{\abs{D^\text{(cf)}}} \sum_{i \in D^\text{(cf)}} \text{acc}_i^\text{(Demo+CF)} \equationPunctuation{,} \\
		& s^\text{(sc)} = \frac{1}{\abs{D^\text{(sc)}}} \sum_{i \in D^\text{(sc)}} \frac{\text{acc}_i^\text{(entail)} + \text{acc}_i^\text{(non-entail)}}{2} \equationPunctuation{,} \\
		& s^\text{(bkd)} = 1 - \frac{1}{\abs{M}\abs{B}} \sum_{i \in B} \sum_{j \in M} \text{ASR}_{ij} \equationPunctuation{.}
	\end{align*}
	\item \emph{Privacy}: $1 - (0.4 \text{LR}^\text{(Enron)} + 0.3\text{LR}^\text{(PII)} + 0.3\text{LR}^\text{(Understand)})$ where
	\begin{align*}
		& \text{LR}^\text{(Enron)} = \frac{1}{T} \sum_{t=1}^T \frac{\text{LR}_t^\text{(Email)} + \text{LR}_t^\text{(Local)} + \text{LR}_t^\text{(Domain)}}{3} \equationPunctuation{,} \\
		& \text{LR}^\text{(PII)} = \frac{1}{P} \sum_{p=1}^P \overline{\text{LR}}^p \equationPunctuation{,} \\
		& \text{LR}^\text{(Understand)} = \frac{1}{WE} \sum_{w=1}^W \sum_{e=1}^E \overline{\text{LR}}_{we} \equationPunctuation{.}
	\end{align*}
	\item \emph{Machine ethics}: $\displaystyle (\text{ACC}^\text{zero} + \text{ACC}^\text{few} + (1 - \overline{\text{FPR}}^\text{jailbreak}) + (1 - \overline{\text{FPR}}^\text{evasive})) / 4 \equationPunctuation{.} $
	\item \emph{Fairness}: $\displaystyle 100 \left( 1 - \frac{M_\text{dpd}^\text{(zero)} + M_\text{dpd}^\text{(few-unfair)} + M_\text{dpd}^\text{(few-fair)}}{3} \right) \equationPunctuation{.}$
\end{itemize}

\begin{wrapfigure}[22]{r}{.5\linewidth}
	\centering
	\vspace{-3.5\baselineskip}
	\includegraphics{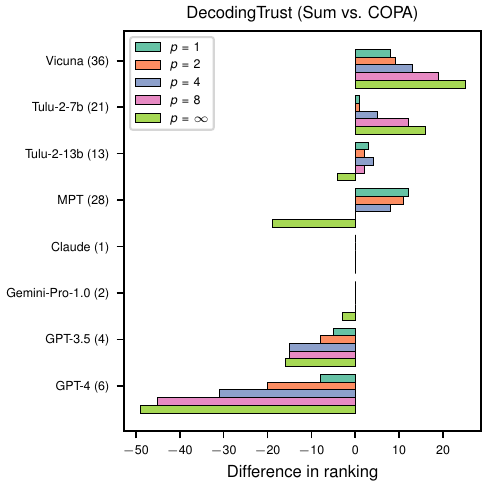}
	\caption{Difference in ranking for a subset of models on the DecodingTrust leaderboard~\citep{wang2023decodingtrust} for different $p$ values of \ours, where the number within parenthesis indicate their original ranking.}%
	\label{app:fig:decodingtrust}
\end{wrapfigure}

\paragraph{Additional results.}

After computing all the variables above, we end up with \num{8} \criteria, which are aggregated in the their official leaderboard by taking the average of all \criteria (\ie, using $p=1$ in \cref{eq:weighted-pnorms} since all values are non-negative).
One natural question then is how do the rankings obtained with the average score change if we plug in \ours on the \criteria given above.
To this end, we take the results published in \href{https://decodingtrust.github.io/leaderboard/}{the official site} of DecodingTrust and recompute their rankings using as aggregated scores \ours with $p \in \{1, 2, 4, 8, \infty\}$.
While this can be easily done in the full leaderboard with \num{55} models to this date (and it is indeed done this way in the code accompanying this manuscript), we show in \cref{app:fig:decodingtrust} a bar plot showing the differences in ranking of a small subset of representative models \wrt average ranking. We summarize the main takeaways as follows:
\begin{enumerate}
	\item \emph{Differences in normalization}: Even with $p=1$, we observe significant differences in \ours \wrt the original ranking where, \eg, Vicuna and MPT rank around 10 positions better and the two GPT models close to 10 worse.
\end{enumerate}

\vspace{-\parskip}%
\begin{enumerate}
\setcounter{enumi}{1}
	\item \emph{Robustness with $p$}: As discussed in the manuscript, increasing (resp. decreasing) $p$ can be interpreted as adding (removing) importance to the performance of the models on individual \criteria. We observe a similar trend here, where those models that lack in one of the specific \criteria (\eg, GPT-4 which performs the worst in the Fairness \criterion) start losing ranks as we increase $p$, and those which are more robust are rewarded instead (\eg, Vicuna and Tulu-2-7b).
	\item \emph{Robustness trend.} Similar to those experiments from \cref{subsec:case-model-analysis}, we observe a clear correlation between the rankings and the values of $p$, that is, the aforementioned robustness criterion is stressed as we increase $p$. While this is not always the case, \eg, Tulu-2-13b fluctuates by one or two rankings as we change $p$, the trend is rather clear.
	\item \emph{Quasi-dominant models}. While there is no a dominant model (\ie, one which is better for all \criteria), some models like Claude and Gemini-Pro-1.0 rank top-10 for all but, respectively, one and two \criteria. As a result, we see that they consistently rank first and second for almost all values of $p$, just like if we took the average. This is, in part, a result of the CDF estimator having less variance for more extreme samples (see \cref{prop:estimator}). In layman's terms, a model that does almost everything great is easy to spot.
	\item \emph{Diversity of solutions.} While not shown in \cref{app:fig:decodingtrust}, it is interesting to remark that in this setting, with \num{8} diverse \criteria, we find that out of the \num{55} available models, the best-worst performing model (that is, the most robust model according to \ours with $p=\infty$) achieves a worst-ranking of \num{22}, with the two second-best LLMs obtaining a worst-ranking of \num{40}. That is, only with \num{8} \criteria we already find that any model performs relatively bad in at least one of them.
\end{enumerate}

\end{document}